\documentclass{article}

\usepackage[preprint]{neurips_2025}

\usepackage[utf8]{inputenc} 
\usepackage[T1]{fontenc}    
\usepackage[colorlinks,linkcolor=blue]{hyperref}
\usepackage{url}            
\usepackage{booktabs}       
\usepackage{amsfonts}       
\usepackage{nicefrac}       
\usepackage{microtype}      
\usepackage[table]{xcolor}  
\definecolor{down}{rgb}{1,0.74901961,0.77254902}
\definecolor{up}{rgb}{0.70196078,0.85882353,0.84705882}
\usepackage{amsmath}
\usepackage{amssymb}
\usepackage{mathtools}
\usepackage{amsthm}
\usepackage{bbm}
\usepackage{algorithm}
\usepackage{algorithmic}
\usepackage{wrapfig}
\usepackage{multirow}
\usepackage{graphicx}
\usepackage{tcolorbox}
\usepackage{pifont}
\usepackage{array}
\usepackage{makecell}
\usepackage{float}
\usepackage{subfig}

\usepackage{enumitem}
\usepackage{hhline}
\usepackage{multirow}

\newcommand{\Tab}{\textbf{Tab.}} 
\newcommand{\Fig}{\textbf{Fig.}}
\newcommand{\Sec}{\textbf{Sec.}}

\title{T2I-ConBench: Text-to-Image Benchmark for Continual Post-training}

\author{
Zhehao Huang$^{1,\dagger}$\quad
Yuhang Liu$^{1,\dagger}$\quad
Yixin Lou$^{1,\dagger}$\quad
Zhengbao He$^{1}$\quad
Mingzhen He$^{1}$
\\
\textbf{%
Wenxing Zhou$^{1}$\quad
Tao Li$^{1}$\quad
Kehan Li$^{2,\ddagger}$\quad
Zeyi Huang$^{2,\mathparagraph}$\quad
Xiaolin Huang$^{1,\mathparagraph}$
}
\\
\\
$^{1}$Shanghai Jiao Tong University\quad
$^{2}$Huawei
\\
\\
\texttt{\textbf{Project Page: \href{https://k1nght.github.io/T2I-ConBench/}{T2I-ConBench}}}
}

\begin{document}

\renewcommand{\thefootnote}{\fnsymbol{footnote}}
\footnotetext[2]{Equal contribution. $^\ddagger$Project leader. $^\mathparagraph$Corresponding authors.}
\renewcommand{\thefootnote}{\arabic{footnote}}

\maketitle

\vspace*{-4mm}
\begin{abstract}
\vspace*{-3mm}
Continual post‑training adapts a single text‑to‑image diffusion model to learn new tasks without incurring the cost of separate models, but naïve post-training causes forgetting of pretrained knowledge and undermines zero‑shot compositionality. We observe that the absence of a standardized evaluation protocol hampers related research for continual post‑training. To address this, we introduce \textbf{T2I‑ConBench}, a unified benchmark for continual post-training of text-to-image models. T2I-ConBench focuses on two practical scenarios, \textit{item customization} and \textit{domain enhancement}, and analyzes four dimensions: (1) retention of generality, (2) target-task performance, (3) catastrophic forgetting, and (4) cross-task generalization. It combines automated metrics, human‑preference modeling, and vision‑language QA for comprehensive assessment. We benchmark ten representative methods across three realistic task sequences and find that no approach excels on all fronts. Even joint “oracle” training does not succeed for every task,
and cross-task generalization remains unsolved. We release all datasets, code, and evaluation tools to accelerate research in continual post‑training for text‑to‑image models.

\end{abstract}

\begin{figure*}[htb]
  \vspace*{-4mm}
  \centering
  \includegraphics[width=0.95\textwidth]{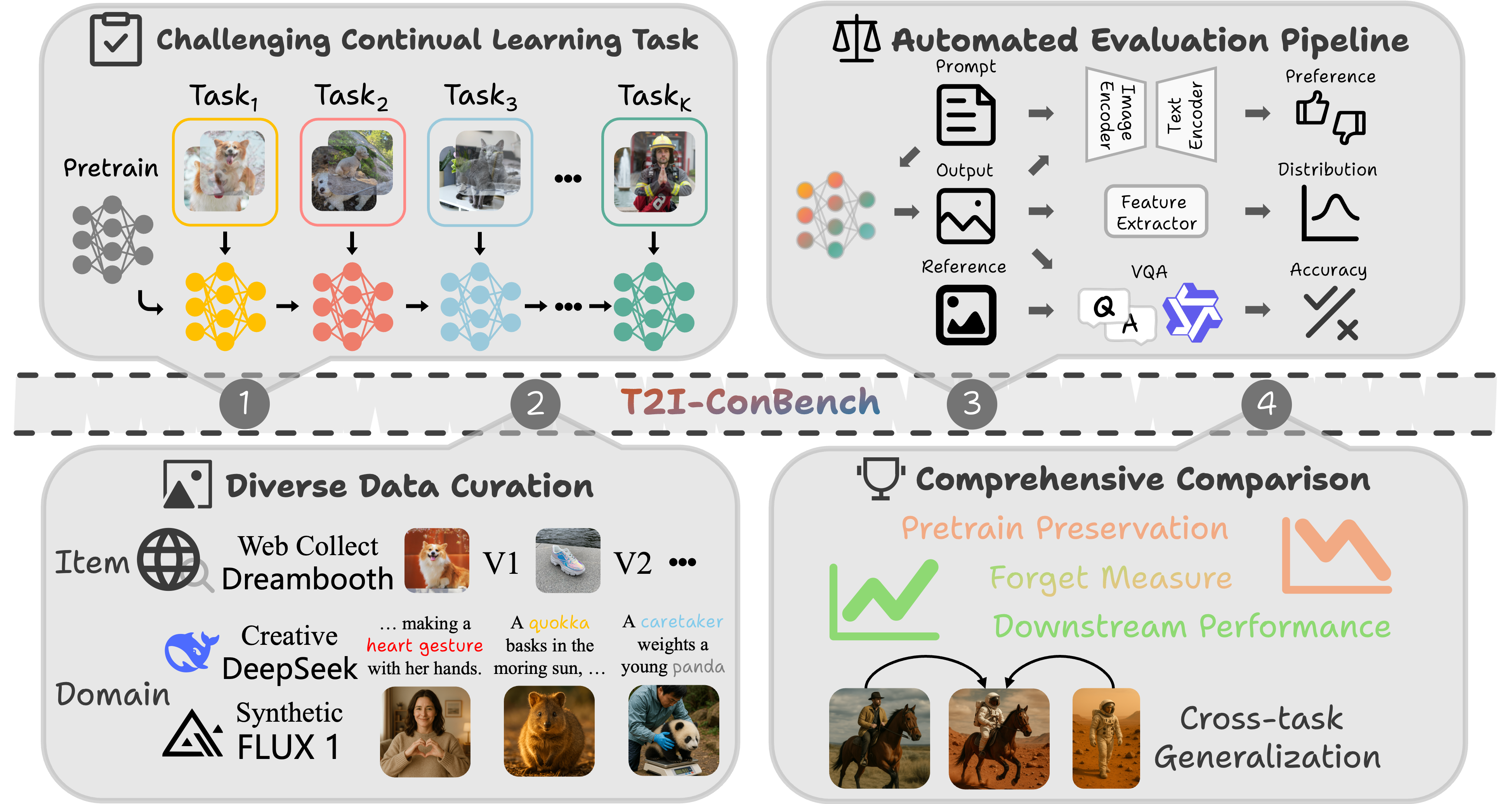}
  \vspace*{-2mm}
  \caption{Overview of T2I-ConBench. Our benchmark consists of four components: (1) challenging continual post‑training task sequences, (2) the curation of diverse item and domain datasets, (3) an automated evaluation pipeline, and (4) comprehensive metrics to fully assess each continual learning method’s ability to update knowledge, resist forgetting, and generalize across tasks.}
  \label{fig:overview}
  \vspace*{-4mm}
\end{figure*}
\section{Introduction}\label{sec:intro}








Over the past few years, large-scale text-to-image (T2I) diffusion models~\cite{DBLP:conf/nips/SahariaCSLWDGLA22,DBLP:conf/cvpr/RombachBLEO22,DBLP:conf/iclr/PodellELBDMPR24,DBLP:conf/iclr/ChenYGYXWK0LL24,DBLP:conf/eccv/ChenGXWYRWLLL24} pretrained on massive image-text corpora have achieved remarkably realistic, high-resolution synthesis. However, real-world deployments~\cite{zhang2025surveypersonalizedcontentsynthesis,DBLP:conf/cvpr/RuizLJPRA23,chaichuk2025promptpolypclinicallyawaremedical,DBLP:conf/siggrapha/KumariS0PSZ24,DBLP:conf/iccvw/CioniBBB23} continually require new concepts, styles, or tasks, ranging from personalized rendering of a specific object to domain‑specific enhancements in medical imaging, industrial design, or cultural heritage. Training and maintaining a dedicated model for each downstream task is impractical due to prohibitive storage overhead and loss of knowledge sharing across tasks~\cite{DBLP:conf/icml/HoulsbyGJMLGAG19,DBLP:conf/iclr/PilaultEP21,DBLP:conf/iclr/HeZMBN22}. 
An ideal solution is to sequentially adapt a single foundation model to each new task dataset, integrating fresh task-specific knowledge while preserving its original pretrained capabilities, commonly referred to as the continual post-training paradigm~\cite{lu2024finetuninglargelanguagemodels,DBLP:journals/tmlr/SmithHZHKSJ24,DBLP:conf/emnlp/KeLS0SL22,DBLP:conf/iclr/KeSLKK023}. 

The key challenge is that, when naively post-trained on new tasks, T2I suffer \textit{catastrophic forgetting}~\cite{DBLP:conf/nips/French93,Ratcliff1990ConnectionistMO}: their ability to generate pretraining concepts degrades as they learn new ones. Recent work~\cite{DBLP:journals/pami/WangZSZ24} has therefore adapted various continual post-training strategies to mitigate this issue, including rehearsal-based methods~\cite{chaudhry2019tinyepisodicmemoriescontinual}, regularization-based methods~\cite{Kirkpatrick_2017,DBLP:conf/icml/ZenkePG17}, and parameter-isolation methods~\cite{DBLP:journals/corr/abs-2404-18466,chen2025mofomomentumfilteredoptimizermitigating,DBLP:journals/tmlr/SmithHZHKSJ24}. They have shown impressive gains in specified scenarios with minimal degradation in general capability. Yet all existing methods evaluate knowledge updates within a single‑granularity, sequential‑task framework and overlook two critical aspects: (1) the dynamic degradation of pretrained capabilities throughout continual adaptation~\cite{DBLP:journals/corr/abs-2404-18466,shi2024continuallearninglargelanguage,wang2023tracecomprehensivebenchmarkcontinual}, and (2) cross‑task generalization~\cite{DBLP:conf/nips/OkawaLDT23,yin2025transformersablereasonconnecting} to combine concepts across tasks. A model subjected to continual downstream learning should not only excel on each new task in isolation, but also preserve its capacity to generalize across both new and previously learned concepts. 
However, there is no unified benchmark to evaluate these trade-offs in continual post-training approaches.

We bridge this gap with \textbf{T2I‑ConBench} (\Fig~\ref{fig:overview}), a comprehensive benchmark for the continual post-training of text-to-image diffusion models. T2I-ConBench covers two prototypical post-training tasks of differing granularity: 
\noindent\ding{182} item customization~\cite{zhang2025surveypersonalizedcontentsynthesis,DBLP:conf/cvpr/RuizLJPRA23}, using web-scraped real-world images to probe personalized object‑level generation, and 
\noindent\ding{183} domain enhancement~\cite{zhu2024domainstudiofinetuningdiffusionmodels}, using synthetic data to test improvement on generative quality and text-image alignment. For each sequence, we craft targeted prompts that challenge both general and specialized generation capabilities. We also develop an automated evaluation pipeline combining standard T2I metrics, a learned human-preference model, and visual question answering to assess 
\noindent\ding{182} preservation of pretrained generality, 
\noindent\ding{183} target‑task performance, 
\noindent\ding{184} forgetting, and 
\noindent\ding{185} cross-task generalization. By unifying these dimensions within one extensible framework, T2I-ConBench enables fair comparison of continual post-training methods, illuminating their relative strengths in updating, retaining, and compositing knowledge.

Building upon T2I-ConBench, we construct three realistic continual post-training scenarios that order tasks of differing granularity, and we evaluate ten representative baseline methods on these mixed-order streams. Our experiments yield three key takeaways: 
\noindent\ding{182} \textit{No single method excels everywhere.} 
\noindent\ding{183} \textit{"Oracle" joint learning is not a panacea.} 
\noindent\ding{184} \textit{Cross-task generalization remains an open challenge.} 

We release all T2I‑ConBench datasets, training scripts, and evaluation pipelines, providing the community with a unified, extensible platform to develop and benchmark continual post‑training strategies for the next generation of T2I diffusion models.
\section{Task Definition}\label{sec:task definition}

\textbf{Continual post-training}~\cite{DBLP:journals/tmlr/SmithHZHKSJ24} of large pretrained T2I diffusion models denotes the sequential adaptation of a single foundation model to a stream of small, task-specific datasets. After each adaptation task, the model must assimilate the novel concepts or domains without access to earlier data and without eroding its original generative competence.
Concretely, we begin with a base model that has completed broad pretraining. We then define a sequence of downstream tasks, each associated with its own disjoint set of text–image pairs. A continual post‑training algorithm produces a new model after each task so that it both adapts to the current task’s data and resists degradation on all previously seen tasks. Achieving this balance requires effective mitigation of catastrophic forgetting while still integrating new knowledge. 
For a more formal definition of tasks, please refer to the Appendix~\ref{sec:appendix task}.
 

\textbf{Cross-task generalization}~\cite{DBLP:conf/nips/OkawaLDT23,yin2025transformersablereasonconnecting} evaluates the ability to recombine knowledge acquired from different tasks into novel concepts. In addition to per‑task performance metrics, our benchmark introduces a compositional generation evaluation to quantify this capability throughout continual post‑training. This ability builds on the key observation that pretrained diffusion models often exhibit zero‑shot generalization~\cite{DBLP:conf/nips/DhariwalN21}, e.g., after learning both “a person riding a horse” and “astronaut” in the pretraining stage, they can generate “an astronaut riding a horse,” which they have never seen during training (\Fig~\ref{fig:overview}). 
We ask: if a model is first continually post‑trained on the “person riding a horse” task and then on the “astronaut” task, does it still retain the ability to produce the novel combination “an astronaut riding a horse”? To answer this, we construct prompts that merge conditions from two different tasks (\Sec~\ref{sec:data}) and then evaluate how reliably the post‑trained model generates images matching these unseen, composed prompts (\Sec~\ref{sec:evaluation}).
By measuring alignment of compositional generations to corresponding prompts, we can determine whether continual post‑training preserves the pretrained model’s generalization to blend concepts. A strong alignment indicates that the continually post-trained model not only learns each task’s concepts but also preserves the representational flexibility to recombine them in novel ways, supporting long‑term accumulation of knowledge.

\textbf{Remark} Unlike traditional T2I benchmarks~\cite{DBLP:conf/nips/HuangSXLL23,DBLP:conf/nips/GhoshHS23,peng2025dreambenchhumanalignedbenchmarkpersonalized} that compare different models, our T2I-ConBench holds both base models and task datasets fixed. We focus on the impact of the continual post-training algorithm itself, without conflating results with variations in data quality or model architecture. Such a design allows us to isolate and precisely measure the impact of continual post-training methods on knowledge retention, downstream performance, and cross-task generalization.

\section{Data Curation}\label{sec:data}

In real-world applications, T2I models often struggle with generating specific items and producing high-quality, domain-specific outputs. Prioritizing only one aspect would leave significant gaps in overall performance. The diverse demands of post-training for T2I models highlight the need for a systematic evaluation framework that accommodates varying data requirements. These data needs can be divided into two main categories:
\begin{itemize}[leftmargin=8pt]
    \item \textbf{Item Customization} focuses on data designed for the personalized generation of specific objects.
    \item \textbf{Domain Enhancement} involves data to improve image quality and semantic consistency within a specific domain (e.g., portrait photography, wildlife images, or natural landscapes).
\end{itemize}


Item Customization and Domain Enhancement differ in granularity and learning objectives, demanding distinct strategies for knowledge updating and retention. These differences imply that the effectiveness of continual post-training methods will depend on task types. These two scenarios form a comprehensive framework for tackling the practical challenges of post-training in T2I models.


For \textbf{Item Customization} tasks, we curate a training dataset comprising four distinct items selected from the dataset provided in \cite{DBLP:conf/cvpr/RuizLJPRA23}. These items are: “V1 dog”, “V2 dog”, “V3 cat”, and “V4 sneaker”\footnote{\url{https://github.com/google/dreambooth}}. The images for these subjects typically capture them under various conditions, environments, and angles to ensure diversity. 
We then use a large language model (LLM) to generate 10 scenarios for each customized item paired with its non-personalized class, forming the \textit{test set for each item}.



\begin{wrapfigure}{r}{0.42\textwidth}
\vspace*{-1.5\baselineskip}        
\begin{center}
    \includegraphics[width=0.95\linewidth]{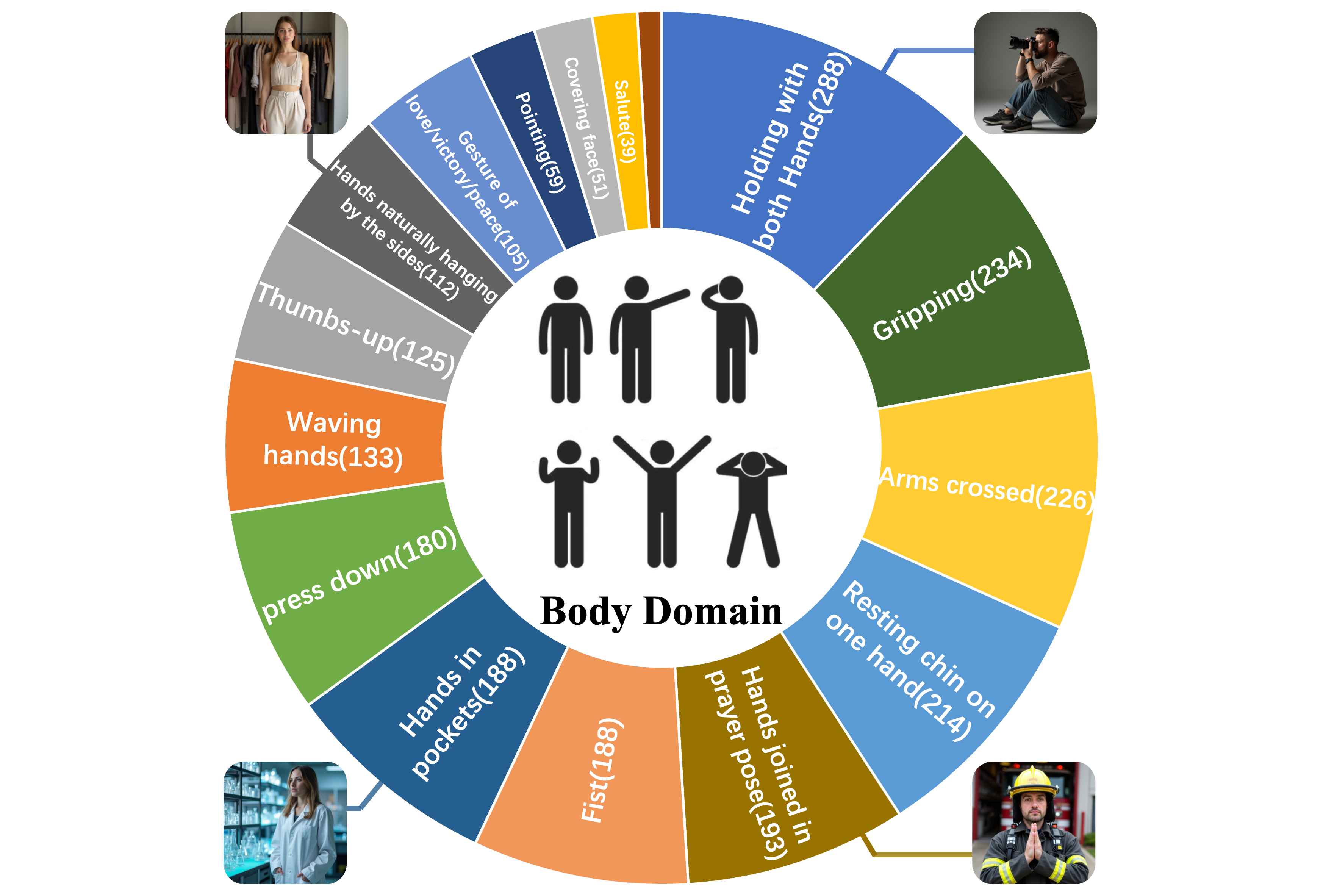}
\end{center}
\vspace*{-3mm}
\caption{\footnotesize{
\textit{Body pose} distribution.
\vspace*{-4mm}
}
}
\label{fig:body_descrip}
\end{wrapfigure}

For \textbf{Domain Enhancement} tasks, we specifically focus on two domains: natural world concepts and human portraits, which we refer to as “\textbf{Nature}” and “\textbf{Body}” domains, respectively. To enhance the base model's image generation quality and semantic alignment within these domains, we first generate numerous prompts containing various concepts within each domain. We then use the base model to test its generation performance on these prompts, identifying concepts where the base model exhibits low generation quality or fails to generate appropriately.
For the “\textbf{Nature}” domain, concepts requiring enhancement include: Squid, Quokka, Markhor, Gerenuk, Spix's Macaw, and Pomelo. For the “\textbf{Body}” domain, we primarily focus on improving the generation of body poses. Concepts requiring enhancement include: pointing, hands naturally hanging by the sides, arms crossed, etc. The total concepts are listed in \Fig~\ref{fig:body_descrip}, along with the number of training data samples for each concept.

To acquire high-quality post-training data for these concepts, we opt for synthetic data generation. Generating synthetic data is an efficient and convenient method for obtaining large, controlled datasets. We first use LLMs to create prompts incorporating the identified concepts. These generated prompts are then sampled; most are designated for the training set, while the remainder form the \textit{test sets for each domain}.
Moreover, to enhance the model's understanding of interactive relationships between concepts across two distinct domains, we construct a training dataset for human interactions with common animals. The training prompts include one common animal concept, for which the base model demonstrates high generation quality, and a concept from the Body domain training set. 
We then use the Flux\_dev model~\cite{flux2024} to generate images for each training-set prompt. The generated data undergo meticulous manual screening to ensure that they are plausible, aesthetically pleasing, and semantically faithful to the prompts. All generated images do not involve any private data and fully comply with established safety and usage standards~\cite{DBLP:journals/bigdatasociety/Beduschi24}.
The initial dataset size is about $80k$. After thorough manual filtering, the final dataset sizes are $2513$ for the Nature domain, $2356$ for body poses, and $1821$ for interactions with common animals. The latter two constitute the Body domain training dataset. For complete information on the dataset, please refer to Appendix~\ref{sec:appendix data}.


\textbf{Cross-Task Generalization Test Sets }
Considering that knowledge across distinct domains is often considered independent, we also aim to investigate the T2I model's generalization capabilities across different domains after continual training. Specifically, we explore the model's ability to synthesize concepts from different domains within a single image. Good generalization capabilities indicate that the model not only learns each task’s concepts but also preserves the representational flexibility to recombine them in novel ways. We construct specialized test sets to probe this cross-dataset generalization:
\begin{itemize}[leftmargin=8pt] 
    \item \textbf{Item+Item:} This set evaluates the model's ability to combine two different trained items in a single image, often within varying environmental contexts. We generate prompts combining pairs of the four trained items within 20 different environmental scenes. 
    \item \textbf{Item+Domain:} These sets evaluate the model's ability to combine a trained item with concepts from either the Nature or Body domains. For the Item-Nature test set, prompts combine each of the five items with various Nature concepts. We generate 3 prompts per item for natural combinations. 
    For the Item-Body test set, prompts combine each of the five items with specific body poses. We generate one prompt for each item-pose pair for a base set of poses, and an additional prompt per item for 11 high-frequency human pose concepts. 
    \item \textbf{Domain+Domain:} To assess the model's ability to combine learned concepts from different domains, we create prompts that combine concepts from the Nature domain training set with concepts from the Body domain training set. This set evaluates if the knowledge learned within distinct domains can be effectively composed when prompted together. For each concept in the Nature domain, its corresponding test set comprises 20 captions, each depicting an interaction between a human and the concept.
\end{itemize}



\section{Evaluation Pipeline}\label{sec:evaluation}

To comprehensively evaluate continual post-training methods, we adopt a multi-axis assessment framework for fair comparison and scalable benchmarking, spanning generation quality, semantic alignment, task-specific accuracy, backward transfer, and compositional generalization.

\textbf{Pretrain Preservation }
To assess how well continual post‑training preserves pretrained capabilities, we use two metrics against the base model.
\noindent\ding{182} \textit{generation quality},
we use Fréchet Inception Distance (\textbf{FID})~\cite{DBLP:conf/nips/HeuselRUNH17} to quantify image‑generation quality, where lower FID indicates closer alignment to real images. We compute FID from the MS-COCO dataset~\cite{DBLP:conf/eccv/LinMBHPRDZ14} as our real‑image reference.
\noindent\ding{183} \textit{text-image alignment}, we employ T2I-CompBench~\cite{DBLP:conf/nips/HuangSXLL23}, which uses a visual language model~\cite{li2025surveystateartlarge,DBLP:conf/iclr/Zhu0SLE24} (VLM) to evaluate the T2I semantic accuracy under compositional prompts. Considering the full T2I‑CompBench involves generating and scoring large, multidimensional datasets, making it costly to run after each task, we select its most representative compositional tasks as a proxy, complex generation (\textbf{Comp}). This subset serves as our metric for post‑training text-image alignment.

\textbf{Downstream Performance }
We define separate evaluation metrics for two downstream tasks with different granularity. 
\noindent\ding{182} \textit{Item Customization}, we measure the model’s accuracy at generating personalized objects. For each fine‑grained concept, we prompt the post‑training model to generate a test set of images, and we use the original concept’s training set of images as references. Employing a designed question prompt template, we then apply a VLM-based visual question answering (VQA)~\cite{Ma2023RobustVQ} pipeline to score the similarity between generated and reference images on the unique personalized concept, denoted as \textbf{Unique‑Sim}.
\noindent\ding{183} \textit{Domain Enhancement}, we assess human aesthetic preference using the Human Preference Score (\textbf{HPS})~\cite{DBLP:conf/iccv/WuSZZL23}, providing a fine‑grained assessment of the aesthetic and semantic fidelity of task domain outputs from T2I models.

\textbf{Forget Measure }
Beyond measuring degradation of pretrained capabilities relative to the base model, we also quantify forgetting in downstream performance dynamics during continual post‑training. For both Item Customization and Domain Enhancement, we compute \textit{backward transfer}~\cite{DBLP:journals/pami/WangZSZ24,DBLP:conf/collas/PradoR22} on their respective downstream metrics, denoted \textbf{Unique‑Forget} and \textbf{Domain‑Forget}. Additionally, we assess forgetting of the base class when learning personalized concepts in Item Customization. We generate images for non-personalized prompts (e.g., "a dog ...") and score their similarity to all personalized examples (e.g., "V1 dog ...") via our VQA pipeline, as \textbf{Class‑Sim}. A lower Class‑Sim indicates less forgetting of the broader class in favor of the specific concept.

\begin{wrapfigure}{r}{0.55\textwidth}
\vspace*{-2\baselineskip}        
\begin{center}
    \includegraphics[width=\linewidth]{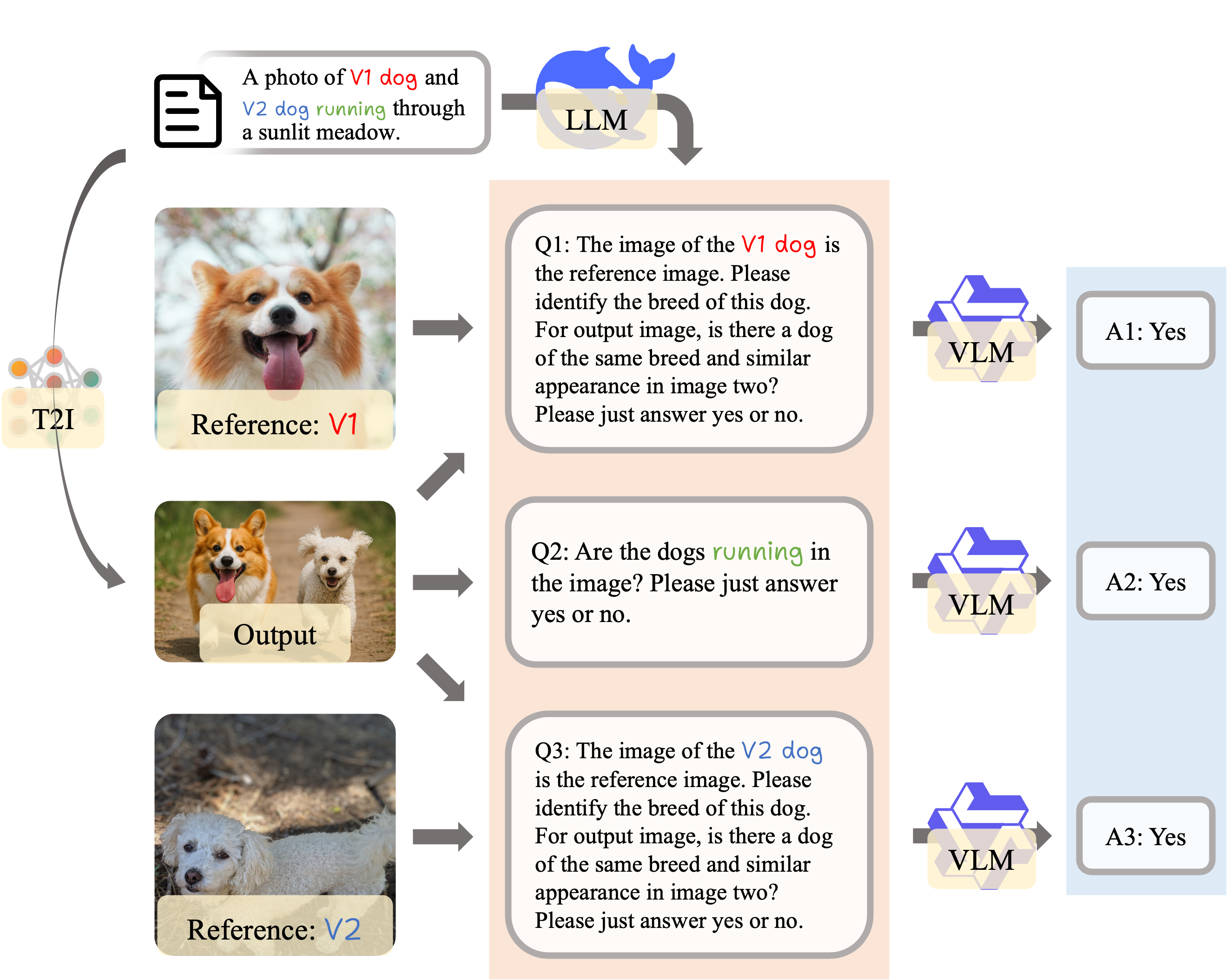}
\end{center}
\vspace*{-4mm}
\caption{\footnotesize{
Evaluation pipeline of cross-task generalization.
\vspace*{-4mm}
}
}
\label{fig:cross_eval}
\end{wrapfigure}
\textbf{Cross-task Generalization } We generate prompts that merge concepts from different tasks and assess whether the fine-tuned model can accurately render these novel combinations. We also score cross‑task performance using a VQA pipeline (\Fig~\ref{fig:cross_eval}). 
First, an LLM decomposes each compositional test‑prompt into its simpler, single‑object components and generates corresponding question‑answer pairs that fully cover both individual object generation and their cross‑task interactions. Next, we convert those Q\&A pairs into VQA‑style questions so that we can directly evaluate image–text alignment by comparing the VLM’s answers against the ground‑truth. For customized item objects or specialized fauna in the nature domain that the VLM may have never seen, we supply reference images of target objects alongside generated images when querying the VLM. Correct responses indicate successful cross-task composition. We evaluate each post-trained model on its respective cross-task test set and report the accuracy as our cross-task generalization metric, reflecting each method’s effects of representational flexibility and long‑term knowledge accumulation. 

\textbf{Remark }
Our evaluation pipeline is fully automated, eliminating the need for human intervention and greatly reducing the labor cost of large‑scale, multi‑round model assessments. The interfaces we define are model‑agnostic, allowing easy integration of more advanced evaluators to improve scoring accuracy. For detailed metric definitions and formulas, please refer to the Appendix~\ref{sec:appendix evaluation}.

\section{Continual Post-training Baselines}\label{sec:baselines}


We refer to the pretrained model as \textbf{Base} for establishing a baseline on general generative capabilities and downstream tasks. We treat the model obtained by jointly training on all task data as the “oracle method”~\cite{DBLP:conf/icassp/WuTPK23}, thereby characterizing the upper bound of performance in sequential learning, as \textbf{Joint}. Specifically for continual post‑training of T2I diffusion models, we apply and adapt 10 baseline methods to mitigate catastrophic forgetting and enhance new concept learning. First, the simplest sequential fine‑tuning (\textbf{SeqFT})~\cite{zhang2024slcaunleashpowersequential,chen2025blip3ofamilyfullyopen} updates all model parameters in task order, optimizing exclusively for the current task without preserving pretrained knowledge or retaining performance on earlier tasks. In addition, we compare the following representative baselines:

\textbf{Rehearsal-based methods} maintain a memory buffer that stores samples to replay prior knowledge. We store 10\% of each completed task’s image–text pairs in the memory buffer and mix them with new‑task data during subsequent post-training. This simple \textbf{Replay} baseline~\cite{chaudhry2019tinyepisodicmemoriescontinual} effectively mitigates forgetting and provides a reference for more advanced rehearsal and buffer‑management strategies.

\begin{figure*}[!t]
  \centering
  \includegraphics[width=\textwidth]{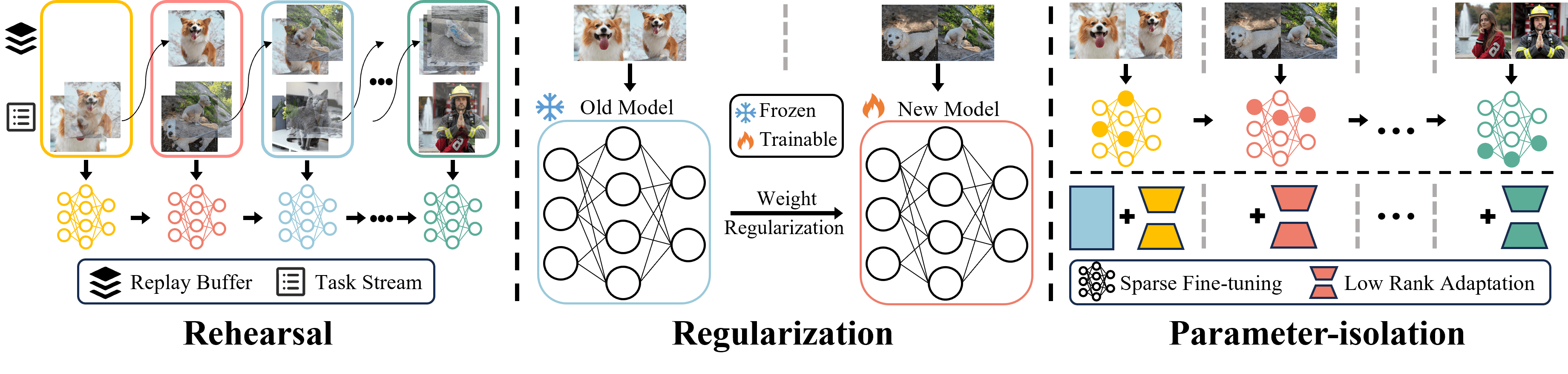}
  \vspace*{-6mm}
  \caption{Overview of the continual post‑training baselines evaluated in this work, encompassing rehearsal‑based, regularization‑based, and parameter‑isolation methods (sparse fine‑tuning and low‑rank adaptation). These baselines are described in \Sec~\ref{sec:baselines} and Appendix~\ref{sec:appendix baselines}.}
  \label{fig:methods}
  \vspace*{-6mm}
\end{figure*}
\textbf{Regularization-based methods} add a constraint term to the training objective to balance between learning new tasks and retaining previous knowledge. We evaluate two regularization baselines: 
\begin{itemize}[left=0pt,itemsep=0pt,topsep=0pt]
    \item \textbf{$\ell_2$-norm}~\cite{DBLP:conf/icml/ZhaoW0L24} adds an $\ell_2$‑norm penalty on the change from the previous task’s final parameters, discouraging significant parameter updates and thus preserving earlier knowledge.
    \item \textbf{EWC}~\cite{Kirkpatrick_2017} weights each parameter’s penalty according to its estimated importance to previous tasks by Fisher information matrix~\cite{liao2018approximatefisherinformationmatrix}. Parameters with higher Fisher scores incur a larger penalty for deviation, thereby more effectively preserving those weights critical to earlier tasks.
\end{itemize}

\textbf{Parameter‑isolation methods} freeze most model parameters and update only a small subset, dramatically reducing the computation and storage costs of full-model post-training. In continual post-training for large T2I diffusion models, they split into two main categories:

\noindent\ding{182} \textbf{Sparse fine‑tuning} updates only a small, sparse subset of parameters, with all others fixed at their initial values. This reduces interference with features learned on previous tasks and mitigates forgetting. We adopt two recently proposed sparse fine‑tuning baselines:
\begin{itemize}[left=0pt,itemsep=0pt,topsep=0pt]
    \item \textbf{HFT}~\cite{DBLP:journals/corr/abs-2404-18466} randomly partitions parameters into two equal groups at each new task. One group (50\%) is trained and the other remains frozen, thereby balancing new concept learning with preservation of prior knowledge.
    \item \textbf{MoFO}~\cite{chen2025mofomomentumfilteredoptimizermitigating} ranks parameters by the absolute value of their Adam momentum after each backward pass, then updates only the top subset for critical directions while freezing the rest. This momentum-driven sparse update efficiently learns new tasks and stabilizes prior performance.
\end{itemize}


\noindent\ding{183} \textbf{Low‑rank adaptation} (LoRA) assumes that the fine-tuning weight update lies in a low‑dimensional subspace. Rather than updating the full weight matrix directly, LoRA factorizes weight changes into the product of two low-rank matrices, while freezing the original weights. This dramatically reduces both storage and computation costs. In the continual post‑training setting, the low‑rank decomposition can be extended into several variants that balance adaptation to new tasks with isolation of prior knowledge. In our experiments, we compare the following four LoRA‑based baselines:
\begin{itemize}[left=0pt,itemsep=0pt,topsep=0pt]
    \item \textbf{SeqLoRA}~\cite{DBLP:conf/naacl/DevlinCLT19} shares a single LoRA adapter across all tasks, updating it cumulatively each round. This approach is simple and efficient, but may suffer from accumulated interference between tasks.
    \item \textbf{IncLoRA}~\cite{DBLP:conf/emnlp/WangCGXBZZGH23} allocates a fresh, independent LoRA adapter for each new task, and sums up all adapters for final inference. By assigning each task its own low‑rank subspace, it enforces strict task isolation at the cost of linearly increasing the number of parameters.
    \item \textbf{O-LoRA}~\cite{DBLP:conf/emnlp/WangCGXBZZGH23} enforces an orthogonality constraint on the up‑projection matrix, making the low‑rank subspaces of different tasks mutually orthogonal.
    \item \textbf{C-LoRA}~\cite{DBLP:journals/tmlr/SmithHZHKSJ24} adds a self‑regularization term that penalizes deviations between the LoRA update for the new task and the adapters learned for previous tasks.
\end{itemize}
\textbf{Remark} For more detailed descriptions of the baselines, please refer to the Appendix~\ref{sec:appendix baselines}. We acknowledge that there are more advanced continual learning techniques~\cite{ren2025learningdynamicsllmfinetuning,peng2025tsvdbridgingtheorypractice} for classification or specialized continual learning methods designed for T2I diffusion models~\cite{DBLP:journals/pami/SunLDLDC24,han2025progressivecompositionalitytexttoimagegenerative}. However, due to the cost of their adaptation and unpredictability to our setup, we do not include them as baselines. The chosen methods are representative and straightforward to illustrate the core properties of each category. In future work, we plan to implement additional approaches to provide further insights.
\section{Experiments}\label{sec:experiments}
\begin{table}[t]
  \caption{Performance of continual post‑training methods on the sequential item‑customization (“V1 dog” $\rightarrow$ “V2 dog” $\rightarrow$ “V3 cat” $\rightarrow$ “V4 sneaker”) and sequential domain enhancement (“Nature” $\rightarrow$ “Body”) task using PixArt‑$\alpha$. $\uparrow$: higher is better. $\downarrow$: lower is better. “I” and “D” denote Item and Domain, with combinations indicating cross‑task generalization evaluations. Excluding \textit{Base} and \textit{Joint}, the best result is in \textbf{bold}, the second‑best is \underline{underlined}. For all metrics except Forget, \colorbox{down}{red cells} indicate a drop of more than 5\% below \textit{Base} for significant degradation, while \colorbox{up}{green cells} indicate an increase of more than 5\% above \textit{Joint} for significant outperformance of the traditional “oracle”.}
  \vspace{3pt}
  \centering
  \small
  \setlength{\tabcolsep}{3pt}
  \resizebox{\textwidth}{!}{
  \begin{tabular}{lcccccc|cccccc}
    \toprule
    \textbf{Order} & \multicolumn{6}{c|}{“V1 dog” $\rightarrow$ “V2 dog” $\rightarrow$ “V3 cat” $\rightarrow$ “V4 sneaker”} & \multicolumn{6}{c}{“Nature” $\rightarrow$ “Body”} \\
    \cmidrule(lr){2-7}\cmidrule(lr){8-13}
    \multirow{2}{*}{\textbf{Method}} &
    \multicolumn{2}{c}{\textbf{Pretrain}} &
    \multicolumn{1}{c}{\textbf{Item}} &
    \multicolumn{1}{c}{\textbf{Cross}} &
    \multicolumn{2}{c|}{\textbf{Forget}} &
    \multicolumn{2}{c}{\textbf{Pretrain}} &
    \multicolumn{2}{c}{\textbf{Domain}} &
    \multicolumn{1}{c}{\textbf{Cross}} &
    \multicolumn{1}{c}{\textbf{Forget}} \\
    \cmidrule(lr){2-3}\cmidrule(lr){4-4}\cmidrule(lr){5-5}\cmidrule(lr){6-7}
    \cmidrule(lr){8-9}\cmidrule(lr){10-10}\cmidrule(lr){11-11}\cmidrule(lr){12-13}
    & \textbf{FID} $\downarrow$ & \textbf{Comp} $\uparrow$ & \textbf{Unique-Sim} $\uparrow$ & \textbf{I+I} $\uparrow$ & \textbf{Class‑Sim} $\downarrow$ & \textbf{Unique‑Forget} $\downarrow$ & \textbf{FID} $\downarrow$ & \textbf{Comp} $\uparrow$ & \textbf{Body‑HPS} $\uparrow$ & \textbf{Nature‑HPS} $\uparrow$ & \textbf{D+D} $\uparrow$& \textbf{Domain‑Forget} $\downarrow$\\
    \midrule
    \textit{Base}         & 26.3153 & 0.3378 & 0.0075 & 0.2250 & 0.0088 & -- 
    & 26.3153 & 0.3378 & 0.2966 & 0.2732 & 0.2637 & -- \\
    \textit{Joint}        & 22.9396 & 0.3308 & 0.2225 & 0.3694 & 0.0695 & -- 
    & 29.0167 & 0.3325 & 0.3032 & 0.2849 & 0.4577 & --  \\
    SeqFT                 & \cellcolor{up} \underline{19.7847} & 0.3319 & 0.2325 & 0.3222 & 0.0633 & 0.8718 
    & \cellcolor{down} 29.9746 & 0.3382 & 0.2939 & 0.2744 & 0.3881 & 0.0392   \\
    \midrule
    SeqLoRA               & 21.9909 & \cellcolor{up} \textbf{0.3493} & 0.0525 &  \underline{0.3500}     & \textbf{0.0263} & 0.6611 
    & \cellcolor{down} 28.4885 & 0.3433 & 0.2997 & 0.2854 & 0.4080 & 0.0083   \\
    IncLoRA               & 21.9657 & 0.3392 & 0.1850 &  0.3278 & 0.0863 & N/A 
    & \cellcolor{down} 28.2885 & \cellcolor{up} \textbf{0.3519} & 0.3006 & 0.2874 & 0.4080 & 0.0007   \\
    O‑LoRA                & 22.6171 & 0.3364 & 0.1775 & 0.2861 & 0.0968 & N/A 
    & \cellcolor{up} 26.5287 & 0.3411 & 0.2942 & 0.2880 & 0.4030 & \underline{-0.0031}  \\
    C‑LoRA                & 23.2204 & 0.3411 & 0.1850 & 0.3056 & 0.0838 & N/A 
    & \cellcolor{up} \textbf{26.1921} & 0.3414 & 0.2920 & \underline{0.2882} & 0.3930 & -0.0031  \\
    \midrule
    $\ell_2$‑norm               & \cellcolor{up} 20.6191 & \underline{0.3417} & 0.1575 & 0.3278 & 0.0468 & 0.7962 
    & \cellcolor{up} 27.1267 & 0.3426 & 0.2990 & 0.2863 & 0.3980 & 0.0003   \\
    EWC                   & \cellcolor{up} 19.8390 & 0.3399 & 0.2250 & 0.3139 & 0.0575 & 0.7017 
    & \cellcolor{down} 29.7816 & 0.3409 & 0.2947 & 0.2746 & 0.3781 & 0.0372   \\
    \midrule
    HFT                   & \cellcolor{up} 20.8671 & 0.3357 & 0.1500 & 0.3028 & \underline{0.0333} & \underline{0.5833} 
    & \cellcolor{down} 28.8833 & 0.3438 & \textbf{0.3010} & 0.2840 & 0.3881 & 0.0104   \\
    MoFO                  & \cellcolor{up} \textbf{19.2802} & 0.3296 & \cellcolor{up} \textbf{0.2850} & 0.3306 & 0.0680 & 0.7296 
    & \cellcolor{down} 29.8326 & 0.3418 & 0.2985 & 0.2803 & \textbf{0.4279} & 0.0196   \\
    \midrule
    Replay                & \cellcolor{up} 20.7805 & 0.3338 & \cellcolor{up} \underline{0.2700} & \textbf{0.3694} & 0.0768 & \textbf{0.1428} 
    & \cellcolor{down} 29.7044 & \cellcolor{up} 0.3508 & \underline{0.3007} & \textbf{0.2890} & \underline{0.4179} & \textbf{-0.0070} \\
    \bottomrule
  \end{tabular}}
  \label{tab:pixart_alpha_item_and_domain}
  \vspace*{-2mm}
\end{table}

\subsection{Implementation Details}


Based on T2I‑ConBench,  we design three continual post-training scenarios for T2I diffusion models with different data granularities: (1) \textit{Sequential item customization} with four fine‑grained concepts learned in order. (2) \textit{Sequential domain enhancement} with two broad domains trained sequentially. (3) \textit{Sequential Item‑Domain Adaptation} with a mixture of the above item and domain tasks, evaluated under two task orders. We evaluate the ten continual post‑training baselines introduced in \Sec~\ref{sec:baselines} on two diffusion architectures, PixArt‑$\alpha$~\cite{DBLP:conf/iclr/ChenYGYXWK0LL24} and Stable Diffusion v1.4~\cite{DBLP:conf/cvpr/RombachBLEO22} (Appendix~\ref{sec:appendix experiment}). Detailed training protocol and hyperparameters are provided in the Appendix~\ref{sec:appendix train}.
\begingroup
\renewcommand{\arraystretch}{0.9}
\begin{table}[t]
  \caption{Performance of continual post‑training methods for the sequential item‑domain adaptation task of two orders using PixArt-$\alpha$. $\uparrow$: higher is better. $\downarrow$: lower is better. “I” and “D” denote Item and Domain, respectively, with combinations indicating cross‑task generalization evaluations. Excluding \textit{Base} and \textit{Joint}, the best result is shown in bold and the second‑best is underlined. For all metrics except Forget, \colorbox{down}{red cells} indicate a drop of more than 5\% below \textit{Base} for significant degradation. Since the traditional “oracle” \textit{Joint} performs poorly in this mixed adaptation scenario, it is not used as the target to surpass.}
  \vspace{3pt}
  \centering
  \resizebox{\textwidth}{!}{
  \begin{tabular}{c|lccccccccc}
    \toprule
    \multirow{2}{*}{\textbf{Order~1}} &
    \multirow{2}{*}{\textbf{Method}} &
    \multicolumn{2}{c}{\textbf{Pretrain}} &
    \multicolumn{1}{c}{\textbf{Item}} &
    \multicolumn{2}{c}{\textbf{Domain}} &
    \multicolumn{3}{c}{\textbf{Cross}} &
    \multicolumn{1}{c}{\textbf{Forget}} \\
    \cmidrule(lr){3-4}\cmidrule(lr){5-5}\cmidrule(lr){6-7}\cmidrule(lr){8-10}\cmidrule(lr){11-11}
    & & \textbf{FID} $\downarrow$ & \textbf{Comp} $\uparrow$ & \textbf{Unique‑Sim} $\uparrow$ & \textbf{Body‑HPS} $\uparrow$ & \textbf{Nature‑HPS} $\uparrow$ &
    \textbf{I+I} $\uparrow$ & \textbf{I+D} $\uparrow$& \textbf{D+D} $\uparrow$ & \textbf{Class‑Sim} $\downarrow$ \\
    \midrule
    \multirow{12}{*}{\makecell[c]{\\“V1 dog”\\$\downarrow$\\“V2 dog”\\$\downarrow$\\“V3 cat”\\$\downarrow$\\“V4 sneaker”\\$\downarrow$\\“Nature”\\$\downarrow$\\“Body”}} &
    \textit{Base}      & 26.3154 & 0.3378 & 0.0075 & 0.2966 & 0.2732 & 0.2250 & 0.3407 & 0.2637 & 0.0088 \\
     & \textit{Joint}     & 29.2236 & 0.3472 & 0.0725 & 0.3054 & 0.2897 & 0.2528 & 0.3898 & 0.4527 & 0.0413 \\
     & SeqFT              & \cellcolor{down} 28.9167 & 0.3483 & 0.0225 & 0.3014 & 0.2832 & 0.2667 & 0.3796 & 0.3980 & 0.0118 \\
    \cmidrule(lr){2-11}
     & SeqLoRA            & \cellcolor{down} 28.7234 & 0.3456 & \cellcolor{down} 0.0000  & 0.3004 & \textbf{0.2890} & 0.2333     & 0.3571 & \underline{0.4129} & N/A      \\
     & IncLoRA            & \cellcolor{down} 28.5758 & 0.3389 & \cellcolor{down} 0.0000  & 0.2965 & 0.2841 & 0.2361   & 0.3919 & 0.3980 & N/A      \\
     & O‑LoRA             & \cellcolor{down} 27.8870 & 0.3388 & 0.0600 & 0.2838 & 0.2838 & \underline{0.2806} & 0.3530 & 0.3632 & \underline{0.0113} \\
     & C‑LoRA             & \textbf{26.5394} & 0.3251 & \underline{0.1175} & 0.2908 & 0.2776 & \textbf{0.2917} & 0.3468 & 0.3085 & 0.0238 \\
    \cmidrule(lr){2-11}
     & $\ell_2$‑norm            & \underline{27.1423} & 0.3425 & 0.0125 & 0.2995 & 0.2860 & 0.2306 & 0.3816 & 0.3930 & \textbf{0.0000} \\
     & EWC                & \cellcolor{down} 28.8256 & 0.3461 & 0.0250 & 0.3016 & 0.2833 & 0.2639 & 0.3877 & \underline{0.4129} & 0.0238 \\
    \cmidrule(lr){2-11}
     & HFT                & \cellcolor{down} 28.8221 & \underline{0.3500} & 0.0375 & \textbf{0.3020} & 0.2827 & 0.2444 & \textbf{0.3918} & 0.3930 & 0.0300 \\
     & MoFO               & \cellcolor{down} 28.8221 & \textbf{0.3500} & 0.0350 & \textbf{0.3020} & 0.2827 & 0.2444 & \textbf{0.3918} & 0.3930 & 0.0300 \\
     \cmidrule(lr){2-11}
     & Replay             & \cellcolor{down} 30.4569 & 0.3461 & \textbf{0.2450} & 0.3006 & \textbf{0.2890} & 0.2556 & 0.3530 & \textbf{0.4527} & 0.0395 \\
    \midrule
    \multirow{2}{*}{\textbf{Order~2}} &
    \multirow{2}{*}{\textbf{Method}} &
    \multicolumn{2}{c}{\textbf{Pretrain}} &
    \multicolumn{1}{c}{\textbf{Item}} &
    \multicolumn{2}{c}{\textbf{Domain}} &
    \multicolumn{3}{c}{\textbf{Cross}} &
    \multicolumn{1}{c}{\textbf{Forget}} \\
    \cmidrule(lr){3-4}\cmidrule(lr){5-5}\cmidrule(lr){6-7}\cmidrule(lr){8-10}\cmidrule(lr){11-11}
    & & \textbf{FID} $\downarrow$ & \textbf{Comp} $\uparrow$ & \textbf{Unique‑Sim} $\uparrow$ & \textbf{Body‑HPS} $\uparrow$ & \textbf{Nature‑HPS} $\uparrow$ &
  \textbf{I+I} $\uparrow$ & \textbf{I+D} $\uparrow$& \textbf{D+D} $\uparrow$ & \textbf{Class‑Sim} $\downarrow$ \\
    \midrule
    \multirow{12}{*}{\makecell[c]{\\“Nature”\\$\downarrow$\\“Body”\\$\downarrow$\\“V1 dog”\\$\downarrow$\\“V2 dog”\\$\downarrow$\\“V3 cat”\\$\downarrow$\\“V4 sneaker”}} &
    \textit{Base}      & 26.3154 & 0.3378 & 0.0075 & 0.2966 & 0.2732 & 0.2250 & 0.3407 & 0.2637 & 0.0088 \\
    & \textit{Joint}     & 29.2236 & 0.3472 & 0.0725 & 0.3054 & 0.2897 & 0.2528 & 0.3898 & 0.4527 & 0.0413 \\
    & SeqFT              & \textbf{19.6193} & 0.3359 & 0.2325 & 0.2950 & 0.3389 & 0.2833 & 0.4430 & 0.3781 & 0.0953 \\
    \cmidrule(lr){2-11}
    & SeqLoRA            & 22.2713 & 0.3433 & 0.1475 & 0.2921 & 0.2723 & \textbf{0.4139} & 0.4430 & 0.3184 & 0.0518 \\
    & IncLoRA            & 23.1411 & \textbf{0.3519} & 0.2300 & 0.2944 & 0.2859 & \underline{0.3889}  & 0.4470 & 0.3433 & 0.2300 \\
    & O‑LoRA             & 22.7191 & 0.3411 & 0.0125 & 0.2862 & 0.2862 & 0.2361 & 0.3632 & 0.3881 & N/A     \\
    & C‑LoRA             & 23.9690 & 0.3414 & 0.0250 & 0.2883 & 0.2867 & 0.2583 & 0.3366 & 0.3781 & N/A     \\
    \cmidrule(lr){2-11}
    & $\ell_2$‑norm            & 20.6750 & 0.3438 & 0.2150 & \textbf{0.3031} & \textbf{0.2912} & 0.3528 & 0.4245 & 0.3831 & 0.0405 \\
    & EWC                & \underline{19.8055} & 0.3449 & \underline{0.2575} & 0.2956 & 0.2775 & 0.3389 & 0.4431 & \underline{0.4229} & 0.0750 \\
    \cmidrule(lr){2-11}
    & HFT                & 22.0834 & 0.3430 & 0.1450 & \underline{0.3023} & 0.2845 & 0.3417 & 0.4368 & 0.4179 & \underline{0.0363} \\
    & MoFO               & 20.5495 & 0.3416 & \textbf{0.3950} & 0.2954 & 0.2783 & 0.3583 & 0.4573 & \textbf{0.4527} & 0.1063 \\
    \cmidrule(lr){2-11}
    & Replay             & \cellcolor{down} 29.0976 & \underline{0.3471} & \cellcolor{down}  0.0000     & 0.3008 & \underline{0.2889} & 0.2389 & 0.3468 & 0.3550 & \textbf{0.0213} \\
    \bottomrule
  \end{tabular}
  }
  \label{tab:pixart_alpha_item_domain_order12}
  \vspace*{-2mm}
\end{table}
\endgroup
\subsection{Continue Post-training for Sequential Item Customization}
The left part of \Tab~\ref{tab:pixart_alpha_item_and_domain} shows PixArt‑$\alpha$’s results on the Sequential Item Customization tasks. All post‑training methods achieve a substantial FID reduction versus the base model, demonstrating that targeted post-training on a small set of high‑quality samples can dramatically boost image fidelity, often called \textit{quality tuning}~\cite{dai2023emuenhancingimagegeneration}. In CompBench’s text-image alignment evaluation, all methods perform roughly on par with the base model. LoRA variants struggle after learning the first task. They typically fail to acquire subsequent concepts, yielding “N/A” for forgetting metrics. This likely reflects LoRA’s constrained update subspace, which cannot span widely differing concepts. Interestingly, SeqLoRA recovers item generation capability when testing on multi-item prompts, yielding an Item+Item generalization score of 0.35. This suggests that SeqLoRA has indeed internalized distinct item concepts, but they only manifest when triggered by specific prompts. Among rehearsal‑free approaches, MoFO performs best, achieving a unique‑item similarity of 28.5\% and lower forgetting than SeqFT. Replay attains 27\% unique‑item similarity and a markedly lower Unique-Sim forgetting (14.28\%), outperforming all rehearsal-free methods and matching Joint in cross-task generalization, benefiting from the scenario’s small dataset sizes. However, despite its efficiency and strong performance, replay may pose privacy risks.

\subsection{Continue Post-training for Sequential Domain Enhancement}


PixArt‑$\alpha$’s performance on the Sequential Domain Enhancement tasks is shown in the right part of \Tab~\ref{tab:pixart_alpha_item_and_domain}. Unlike in item customization, the results of most methods get increased FID and indicate a degradation in overall image quality. The underlying reason is that fine‑tuning directly on the new domain erodes the model’s coverage of the general image distribution. Nonetheless, all methods achieve modest gains on CompBench, indicating improved text-image alignment with the target domain.
LoRA variants perform well at domain learning. They yield strong human preference scores, even outperforming Joint on the Nature domain, and exhibit low domain forgetting. Yet they struggle to capture the more complex variations in the body domain, limiting their gains there.
HFT achieves the highest HPS on the Body domain. Its strategy of reusing half the parameters and features effectively learns the detailed motions characteristic of body images.
Replay remains the top performer on downstream metrics, and even achieves positive backward transfer (–0.70\% Domain-Forget), implying that shared domain features can reinforce earlier knowledge. Exploring how to exploit these commonalities for more effective continual updating is a promising direction.
MoFO delivers the best cross‑task generalization (42.79\%), though it is still behind Joint by 2.98\%.

\subsection{Continue Post-training for Sequential Item-Domain Adaptation}


The results for the Item‑Domain Adaptation setting are reported in \Tab~\ref{tab:pixart_alpha_item_domain_order12}, corresponding  to the two task orders we investigate: \textit{Order~1} learns items first, then domains, and \textit{Order~2} learns domains first, then items. Because the item and domain datasets differ substantially in size and quality, this imbalance will induce a pronounced effect on continual learning.

In both task orders, \textit{pretraining preservation} follows the second task: when domain enhancement follows item customization (Order~1), all methods see FID increase as in \Tab~\ref{tab:pixart_alpha_item_and_domain}, mirroring the degraded image quality observed in sequential domain enhancement. Conversely, when item customization comes second (Order~2), FID decreases during that stage. Across both orders, CompBench scores improve for nearly every method, demonstrating consistent gains in text-image alignment through continual post-training.
For \textit{downstream performance}, LoRA variants split into unregularized (SeqLoRA, IncLoRA) and regularized (O-LoRA, C-LoRA) groups. The unregularized methods completely forget items in Order~1, yielding 0.0 accuracy. By contrast, the regularized methods preserve item accuracy when items are learned first but degrade significantly in Order~2, indicating that domain-task regularization can interfere with later item adaptation. Other regularization and sparse fine-tuning techniques also achieve strong results on whichever task is learned second, yet suffer severe forgetting on the first task. For example, unique-item accuracy for initially learned items nearly drops to zero in Order~1.
Replay behaves differently from all others across the two orders. Its performance on the domain‑enhancement task is insensitive to task order, but it only excels when items are learned first. When items come second, Replay fails to acquire the new item‑specific features. We hypothesize that, in Order~2, replaying the larger domain dataset severely interferes with learning the minority specialized item concepts.
Notably, Joint also struggles in this imbalanced data‑stream setup. Dominated by the larger domain dataset, Joint effectively overfits to domain enhancement and fails to learn the fine-grained personalized generation required for items.

For \textit{cross‑task generalization}, Joint also loses the benefits of separately training on items and domains in both orders. Because it underfits the item tasks, Joint performs poorly on Item$+$Item and Item$+$Domain generalization, though it remains best on Domain$+$Domain. The LoRA variants are primarily driven by their performance on item tasks. C-LoRA and O-LoRA achieve the highest Item$+$Item metrics in Order~1 but collapse in Order~2. Conversely, SeqLoRA and IncLoRA reverse that trend. All four LoRA methods exhibit weak cross-task generalization when paired with domain tasks.
Regularization methods ($\ell_2$‑norm, EWC) and sparse fine‑tuning methods (HFT, MoFO) perform poorly under Order~1 but nearly match or exceed Joint in Order~2. This indicates that task sequence not only affects knowledge updating and forgetting, but also the fusion and generalization of learned concepts. 
Finally, Replay fails to balance rehearsal of old data with adaptation to new data, resulting in weak cross‑task generalization under both orders.
Crucially, continual post‑training sequences that first reinforce the coarse‑grained domain and then learn fine‑grained items emerge as a particularly promising direction.

\subsection{Results Summary}

Summarizing the experimental results across the three settings, we draw three key takeaways:

\noindent\ding{182} \textit{No single method excels everywhere.}
Although LoRA variants indeed minimize forgetting, it severely degrades performance on item customization. Other rehearsal‑free methods learn and preserve more knowledge than SeqFT, yet they still exhibit varying degrees of forgetting. Replay performs well under balanced data streams but its effectiveness becomes unstable under imbalanced streams. These results motivate the development of advanced continual post-training methods for T2I diffusion models that better reconcile the trade-off between stability and plasticity.


\noindent\ding{183} \textit{“Oracle” Joint learning is not a panacea.}
In classical continual learning, Joint learning on all datasets is typically treated as the “oracle” upper bound. However, our study reveals that, although Joint usually outperforms baseline continual post‑training methods in most scenarios, it can struggle conflicting demands of multi-task optimization, failing to reach optimal performance on specific domains, a limitation also observed in prior work~\cite{chen2025blip3ofamilyfullyopen}. Furthermore, under imbalanced tasks, Joint often overlooks few-shot concepts, such as minority items. These findings underscore both the challenge posed by our benchmark and the promising solution of continual post‑training.


\noindent\ding{184} \textit{Cross‑task generalization remains an open challenge.}
In both the sequential item customization and domain enhancement scenarios, most methods fall short of Joint in cross‑task generalization. Although many baselines can alleviate catastrophic forgetting, few match the oracle’s ability to seamlessly recombine prior and newly acquired knowledge. This gap highlights the need for approaches that not only preserve prior representations but also actively integrate them with incoming information. For example, identifying shared parameters and features that can be reused to bootstrap new-task learning offers a promising path to enhance cross-task generalization. To accelerate this progress, we provide a standardized evaluation protocol within T2I-ConBench, empowering the continual learning community to develop and rigorously benchmark more sophisticated post-training methods.

\section{Conclusions}

This paper presents T2I-ConBench, a comprehensive benchmark for continual post-training of T2I diffusion models. We curate datasets spanning open-world scenarios with two levels of granularity and develop an automated evaluation pipeline that measures preservation of pretrained capabilities, downstream performance, forgetting, and cross-task generalization. We evaluate and analyze representative continual post-training methods across three sequential-task settings, establishing comparative baselines and insights to guide the development of more advanced methods. We hope that T2I-ConBench could serve as a standardized testing framework to accelerate both research and practical deployment of continual post-training techniques for T2I diffusion models.

\newpage
{\small
\bibliographystyle{unsrt}
\bibliography{ref}
}


\newpage
\appendix

\onecolumn
\setcounter{section}{0}

\section*{Appendix}
\setcounter{section}{0}
\setcounter{figure}{0}
\setcounter{prop}{0}
\makeatletter 
\renewcommand{\thefigure}{A\arabic{figure}}
\renewcommand{\theHfigure}{A\arabic{figure}}
\renewcommand{\thetable}{A\arabic{table}}
\renewcommand{\theHtable}{A\arabic{table}}

\makeatother
\setcounter{table}{0}

\setcounter{algorithm}{0}
\renewcommand{\thealgorithm}{A\arabic{algorithm}}
\setcounter{equation}{0}
\renewcommand{\theequation}{A\arabic{equation}}

\section{Related Work}\label{sec:related work}

\subsection{Large-scale Text-to-image Generative Model}

Large‑scale text‑to‑image (T2I) diffusion models have rapidly become the backbone of generative AI. Building on latent diffusion, Stable Diffusion~\cite{DBLP:conf/cvpr/RombachBLEO22,DBLP:conf/iclr/PodellELBDMPR24} popularized an open‑source U‑Net~\cite{ronneberger2015unetconvolutionalnetworksbiomedical} conditioned on CLIP~\cite{radford2021learningtransferablevisualmodels}, capable of efficient generation by operating in a compressed latent space. Meanwhile, the PixArt series~\cite{DBLP:conf/iclr/ChenYGYXWK0LL24,DBLP:conf/eccv/ChenGXWYRWLLL24} demonstrates that decomposed training stages, latent consistency modules, and weak‑to‑strong paradigms can reduce training cost by over 90\%, while supporting 4K output and 2–4‑step sampling for sub‑second inference. The latest \href{https://github.com/black-forest-labs/flux}{FLUX.1} models from Black Forest Labs scale diffusion transformers to 12B parameters with spatiotemporal attention and multi‑stage noise scheduling, matching \href{https://www.midjourney.com/home}{Midjourney} and DALL~E3~\cite{shi2020improvingimagecaptioningbetter} in fidelity and prompt adherence. Crucially, pre‑training on diverse, large‑scale image–text data endows these models with strong zero‑shot generalization, enabling them to adapt to downstream domain‑specific or personalized tasks with minimal post‑training.

\subsection{Continual Post-training for Image Generation}

Continual post-training~\cite{lu2024finetuninglargelanguagemodels,DBLP:journals/tmlr/SmithHZHKSJ24,DBLP:conf/emnlp/KeLS0SL22,DBLP:conf/iclr/KeSLKK023} enables a single, large T2I diffusion model to absorb new, task-specific knowledge without full retraining, yielding substantial improvements on practical downstream applications. We target two key scenarios: \textbf{item customization}~\cite{zhang2025surveypersonalizedcontentsynthesis}, where the model must learn to generate a novel object or style from only a few examples while maintaining consistency across diverse contexts, and \textbf{domain enhancement}~\cite{guo2024syntheticdiffusiondriventesttimeadaptation}, which focuses on refining overall image quality and semantic fidelity within a specialized visual domain. In item customization, methods such as C-LoRA~\cite{DBLP:journals/tmlr/SmithHZHKSJ24} incrementally inject new concepts into cross-attention layers via low-rank adapters, while regularizing against forgetting; encoder-based adapters learn a compact network that maps reference images into embeddings fused into the diffusion process for rapid personalization~\cite{su2023identityencoderpersonalizeddiffusion}; and even zero-training approaches repurpose attention maps from exemplars at inference time to steer generation without further optimization~\cite{hertz2022prompttopromptimageeditingcross}. For domain enhancement, techniques like Diffuse-UDA~\cite{gong2024diffuseudaaddressingunsuperviseddomain} and DiffBoost~\cite{zhang2024diffboostenhancingmedicalimage} adapt diffusion priors to medical imaging by aligning appearance and structural statistics or leveraging expert-model features, achieving high-fidelity lesion synthesis and enhanced segmentation generalization. Similarly, portrait-specific fine-tuning and 3D-aware adapter schemes improve face generation fidelity and multi-view consistency~\cite{gu2024diffportrait3dcontrollablediffusionzeroshot,wei2024highfidelity3dportraitgeneration}. Although these approaches deliver strong results in their respective settings, they focus on isolated, single-granularity task sequences and do not evaluate a model’s capacity to recombine concepts across different domains. To address this gap, we propose a unified, sequential benchmark that integrates both item customization and domain enhancement, challenging models to preserve their pretrained versatility, master new tasks, and sustain multi-domain knowledge generalization.

\subsection{Benchmarking Image Generation}

Benchmarking image‑generation models requires a suite of metrics that capture quality, diversity, and alignment with text prompts. Inception Score (IS)~\cite{barratt2018noteinceptionscore} evaluates sharpness and diversity by measuring the confidence and entropy of class predictions from a pretrained Inception‑v3~\cite{DBLP:conf/cvpr/SzegedyVISW16} network on generated samples. The Fréchet Inception Distance (FID)~\cite{DBLP:conf/nips/HeuselRUNH17} compares the mean and covariance of deep Inception features between generated and real images, quantifying distributional similarity. To assess perceptual similarity, LPIPS~\cite{zhang2018unreasonableeffectivenessdeepfeatures}, CLIP‑I~\cite{radford2021learningtransferablevisualmodels}, and DINO Score~\cite{caron2021emergingpropertiesselfsupervisedvision} compute distances in learned feature spaces, reflecting human judgments of visual similarity. Global text–image alignment is measured by multimodal encoders via CLIP‑T, CLIPScore~\cite{radford2021learningtransferablevisualmodels}, and BLIP~\cite{li2022blipbootstrappinglanguageimagepretraining}, which score how well an image matches its prompt in the joint embedding space. For fine‑grained semantic and logical fidelity under complex prompts, benchmarks like GenEval~\cite{DBLP:conf/nips/GhoshHS23} using object detectors and T2I‑CompBench~\cite{DBLP:conf/nips/HuangSXLL23} probe category‑ and relation‑level understanding. To capture human preference, learned reward models such as HPS~\cite{DBLP:conf/iccv/WuSZZL23} and ImageReward~\cite{xu2023imagerewardlearningevaluatinghuman} encode crowd‑sourced judgments into automatic scores. Recent personalization benchmarks, DreamBench~\cite{DBLP:conf/cvpr/RuizLJPRA23} and DreamBench++~\cite{peng2025dreambenchhumanalignedbenchmarkpersonalized}, leverage multimodal LLMs~\cite{li2025surveystateartlarge,DBLP:conf/iclr/Zhu0SLE24} to evaluate object‑level customization quality. Building on these, we introduce a vision‑language‑LLM‑QA‑based pipeline~\cite{Ma2023RobustVQ} to measure cross‑task generalization, which is the ability to recombine old and new concepts across sequential downstream tasks. By extending static metrics into dynamic continual‑learning streams, our benchmark quantifies not only per‑task performance and forgetting but also knowledge transfer and synergy between tasks. CLoG~\cite{zhang2024clogbenchmarkingcontinuallearning} also aims at benchmarking continual learning of generative models, but unlike its continual pre‑training setting starting from scratch, we focus on continual post‑training, and uniquely assesses both retention of pre‑trained zero‑shot capabilities and knowledge generalization in mixed‑task streams.

\section{Broader Impact and Limitations}\label{sec:appendix impact limit}

\textbf{Impact}
T2I-ConBench fills a critical gap in continual post-training evaluation by introducing, for the first time, a unified protocol that measures pretrained capability preservation, downstream task performance, catastrophic forgetting, and cross-task compositional generalization—laying a solid foundation for fair comparisons and reproducible research. Through two systematic tasks—“item customization” and “domain enhancement”—the benchmark not only uncovers the key trade-offs between retaining prior knowledge and adapting to new tasks, but also quantifies the dynamic degradation of old and new concept performance and the shortcomings of zero-shot composition. By releasing our datasets, prompt libraries, and evaluation pipeline, we dramatically lower the barrier for both research and deployment, spurring innovation across diverse application domains such as industrial design, medical imaging, and cultural heritage. At the same time, we must remain vigilant about potential downsides: a standardized continual post-training toolkit could be misused to rapidly produce highly realistic deepfakes or personalized attacks, heightening privacy risks and misinformation. Moreover, the automated evaluation metrics and pretrained models underpinning our benchmark may carry social biases, risking the inadvertent perpetuation or amplification of unfairness.

\textbf{Limitation}
Despite its unique breadth of evaluation, T2I-ConBench has several limitations. First, for precise concept-targeted generation, we rely on the FLUX model, meaning our synthetic data may inherit its biases and constraints in detail fidelity and aesthetic style, which can limit our capacity to fully assess semantic accuracy and visual consistency. Second, we focus exclusively on diffusion architectures and omit equally popular autoregressive generative models, whose differing training regimes and inductive biases could affect the relative performance of various continual post-training methods—an open question for future study. Finally, due to computational resource constraints, we evaluated on stable, mid-scale models rather than the largest and most cutting-edge networks. Nevertheless, our evaluation pipeline is model-agnostic and can readily incorporate the latest diffusion or autoregressive models going forward.

\section{Detailed Task Definition}\label{sec:appendix task}

\textbf{Continual post-training} of large pre‑trained T2I diffusion models refers to the process of sequentially fine‑tuning a single foundation model on a series of small, task‑specific datasets. The models are expected to fine-tune on each task without revisiting earlier data to customize to new domains or concepts while preserving their original generative capabilities. Concretely, let a model $M_0$ with parameters $\theta_0$ have completed a broad pre-training task $\mathcal T_0$. We then define a downstream task sequence $\{\mathcal T_1, \mathcal T_2,\dots, \mathcal T_K\}$. Each task $\mathcal T_i$ provides a dataset $\mathcal D_i=\{(x_{i,n},y_{i,n})\}_{n=1}^{N_i},1\leq i\leq K$ of $N_i$ text-image pairs sampled from distribution $P_{x_i,y_i}$. The datasets are disjoint, $\mathcal D_{i}\cap \mathcal D_j = \empty,i\neq j$. A continual post-training algorithm $\mathcal A$ produces a sequence of models $M_i=\mathcal A(M_{i-1},\mathcal D_i)$ such that each $M_i$ both maximizes the likelihood $p_{M_i}(\hat y|x_i)$ on new task $\mathcal T_i$ and minimizes degradation on all previous tasks $\{\mathcal T_0,\mathcal T_1,\dots,\mathcal T_{i-1}\}$. Balancing these objectives requires effective mitigation of catastrophic forgetting while still integrating new knowledge. Unlike traditional benchmarks on image generation that compare different models or training datasets, our continual post-training benchmark fixes both the base model and the task datasets. It is therefore a systematic evaluation of continual learning strategies: isolating the impact of training algorithms, without conflating results with variations in data quality or model architectures. This design enables precise measurement of how different continual post-training methods truly affect downstream performance, preservation of prior knowledge, and cross-task generalization.
 

\textbf{Cross-task generalization} evaluates the ability to recombine knowledge acquired from different tasks into novel concepts. In addition to per‑task performance metrics, our benchmark introduces a compositional generation evaluation to quantify this property during continual post‑training. This builds on the key observation that pre‑trained diffusion models often exhibit zero‑shot compositionality, e.g., after learning both “a person riding a horse” and “astronaut” in the pre-training stage, they can generate “an astronaut riding a horse,” which they have never seen during the training process. We wonder whether, when a model is continually post-trained first on $\mathcal T_1$ ("a person riding a horse") and then on $\mathcal T_2$ (“astronaut”), does it retain the ability to produce the novel concept $\mathcal T_1\cup \mathcal T_2$ (“an astronaut riding a horse”)? Formally, let $g(x_i,x_j)$ be a semantic-composition function that combines two prompt conditions from tasks $\mathcal T_i$ and $\mathcal T_j$. After obtaining the model $M_i$ via continual post-training on tasks $\{\mathcal T_0,\dots,\mathcal T_i\}$, we measure its cross-task generalization by conditional generation likelihood $p_{M_i}(\hat y|g(x_i,x_j))$ for pairs $(x_i,x_j)$ drawn from different tasks. A high generation likelihood indicates that the model not only learned each task’s concepts but also preserved the representational flexibility to recombine them in novel ways. This metric thus reveals whether continual post‑training sustains the emergent compositional structure of pre‑trained knowledge and supports long‑term accumulation of generative capabilities.
\section{Detailed Dataset Description}\label{sec:appendix data}

\textbf{Dataset Curation Process } involves Identifying Challenging Concepts, Prompt Creation, Image Generation, and Quality Filtering, details are given below:

\noindent\ding{182} \textit{Identifying Challenging Concepts }
For the construction of our domain-specific datasets, we specifically targeted concepts within the chosen domain that the base model either failed to generate entirely or rendered with low quality. The initial step involved identifying these 'challenging concepts.' This was achieved by prompting the base model to generate images for a wide array of domain-relevant concepts, followed by a manual visual screening of the results to pinpoint specific concepts requiring quality improvement.

\noindent\ding{183} \textit{Prompt Creation }
Once the challenging concepts were identified, we utilized a LLM to construct a diverse set of descriptive captions featuring these specific concepts. This collection of captions was subsequently divided through random sampling to serve distinct purposes. The majority of these captions were allocated as prompts for the training dataset, while the remaining smaller portion was set aside to form the test set, intended for later evaluation of model capabilities within this domain.

\noindent\ding{184} \textit{Image Generation and Quality Filtering }
Critically, this image generation step utilized the previously allocated training prompts with a higher-fidelity text-to-image model, chosen specifically for its superior generation quality over the base model. However, these generated images were not used directly. They first underwent a meticulous manual filtering process, where evaluators carefully screened each image for relevance to the prompt, visual quality, and overall coherence.

The final dataset sizes are $2513$ for the Nature domain, $2356$ for body poses, and $1821$ for interactions with common animals. The latter two constitute the Body domain training dataset. The detailed information of the domain-enhancement dataset is shown in \Tab~\ref{tab:append_data_table}.


\begin{table}[htbp]
\centering
\caption{The detailed concepts of the training dataset.}\label{tab:append_data_table}
{
\begin{tabular}{cccc}
\toprule
Category & Specific Actions or Objects & Count & Total \\ \midrule
\multirow{5}{*}{Nature}
 & Gerenuk & 1043 & \\
 & Spix's Macaw & 590 & \\
 & Quokka & 492 & 2513\\
 & Pomelo & 363 & \\ 
 & Squid & 25 & \\
 \midrule
\multirow{16}{*}{Body: Poses} 
 & Hands naturally hanging by the sides & 112 &\multirow{16}{*}{2356} \\
 & Gestures of hearts, victory, peace & 105 & \\
 & Hands joined  in prayer pose & 193 &  \\
 & Resting chin on one hand & 214 & \\
 & Holding with both hands & 288 & \\
 & Hands in pockets & 188 & \\
 & Covering Face & 51 & \\
 & Waving hands & 133 & \\
 & Arms crossed & 226 & \\
 & Thumbs-up & 125 & \\
 & Press down & 180 & \\
 & Gripping & 234 & \\
 & Pointing & 59 & \\
 & Salute & 39 & \\
 & Fist & 188 & \\
 & Others & 21 & \\ 
 \midrule
\multirow{13}{*}{Body: Interaction with} & Dog & 247 & \\
 & Elephant & 138 &\multirow{13}{*}{1821} \\
 & Panda & 177 & \\
 & Tiger & 135 & \\
 & Cat & 192 & \\
 & Monkey & 136 &  \\
 & Horse & 27 & \\
Common Animals & Butterfly & 189 & \\
 & Lion & 173 & \\
 & Giraffe & 111 & \\
 & Dolphin & 88 & \\
 & Kangaroo & 185 & \\
 & Penguin & 23 & \\ 
 \bottomrule
\end{tabular}
}
\end{table}
\section{Detailed Evaluation Pipeline}\label{sec:appendix evaluation}
Benchmarking continual learning methods requires not only the evaluation of static tasks, but also the dynamic evaluation of the performance of the text graph model to detect the performance improvement of downstream tasks and the forgetting of old task knowledge. We designed a unified indicator selection for evaluating the quality of different aspects of text graphs. In addition to the final performance and forgetting metrics commonly used in continuous learning benchmarks, we also focus on the changes in the general capabilities of large models, measure model performance from two aspects: generation quality and semantic logic, and pay attention to the evaluation of cross-task generalization capabilities.

\textbf{Pretrain Preservation}
To assess how well continual post‑training preserves pretrained capabilities, we use two metrics against the base model.

\noindent\ding{182} \textit{Generation Quality}
We use Fréchet Inception Distance (\textbf{FID})~\cite{DBLP:conf/nips/HeuselRUNH17} to evaluate both the quality and diversity of images generated by diffusion models. By comparing the statistical distributions of generated versus real images in the feature space of a pretrained network, FID quantifies how closely a model’s output matches the true data distribution. We use fixed 30,000 captions from MS-COCO~\cite{DBLP:conf/eccv/LinMBHPRDZ14} to generate images for measuring the quality of T2I models. The lower the FID value, the better the quality of the generated images indicated by the trained model. We implement FID from \href{https://github.com/boomb0om/text2image-benchmark}{text2image-benchmark} and employ \href{https://github.com/mseitzer/pytorch-fid/releases/download/fid_weights/pt_inception-2015-12-05-6726825d.pth}{Inception V3} model~\cite{DBLP:conf/cvpr/SzegedyVISW16} as the pretrained network with \href{https://github.com/boomb0om/text2image-benchmark/releases/download/v0.0.1/MS-COCO_val2014_fid_stats.npz}{precomputed FID stats}.

\noindent\ding{183} \textit{Text-image Alignment} \href{https://github.com/Karine-Huang/T2I-CompBench}{T2I-CompBench}~\cite{DBLP:conf/nips/HuangSXLL23} is a comprehensive benchmark for open-world combinatorial T2I generation, providing a large-scale dataset and semantic logic and text-image alignment evaluation metrics. The evaluation dimensions include: multi-entity relationship construction, precise attribute binding, spatial reasoning consistency, and cross-modal semantic fidelity. Through multimodal large language model evaluation, the semantic accuracy of text-to-image models under complex text prompts can be evaluated. We select the 3-in-1 evaluation for complex compositions (\textbf{Comp}) as our metrics.

\textbf{Downstream Performance} We define separate evaluation metrics for two downstream tasks with different granularity. 

\begin{figure}[!htp]
    \centering
    \includegraphics[width=0.7\textwidth]{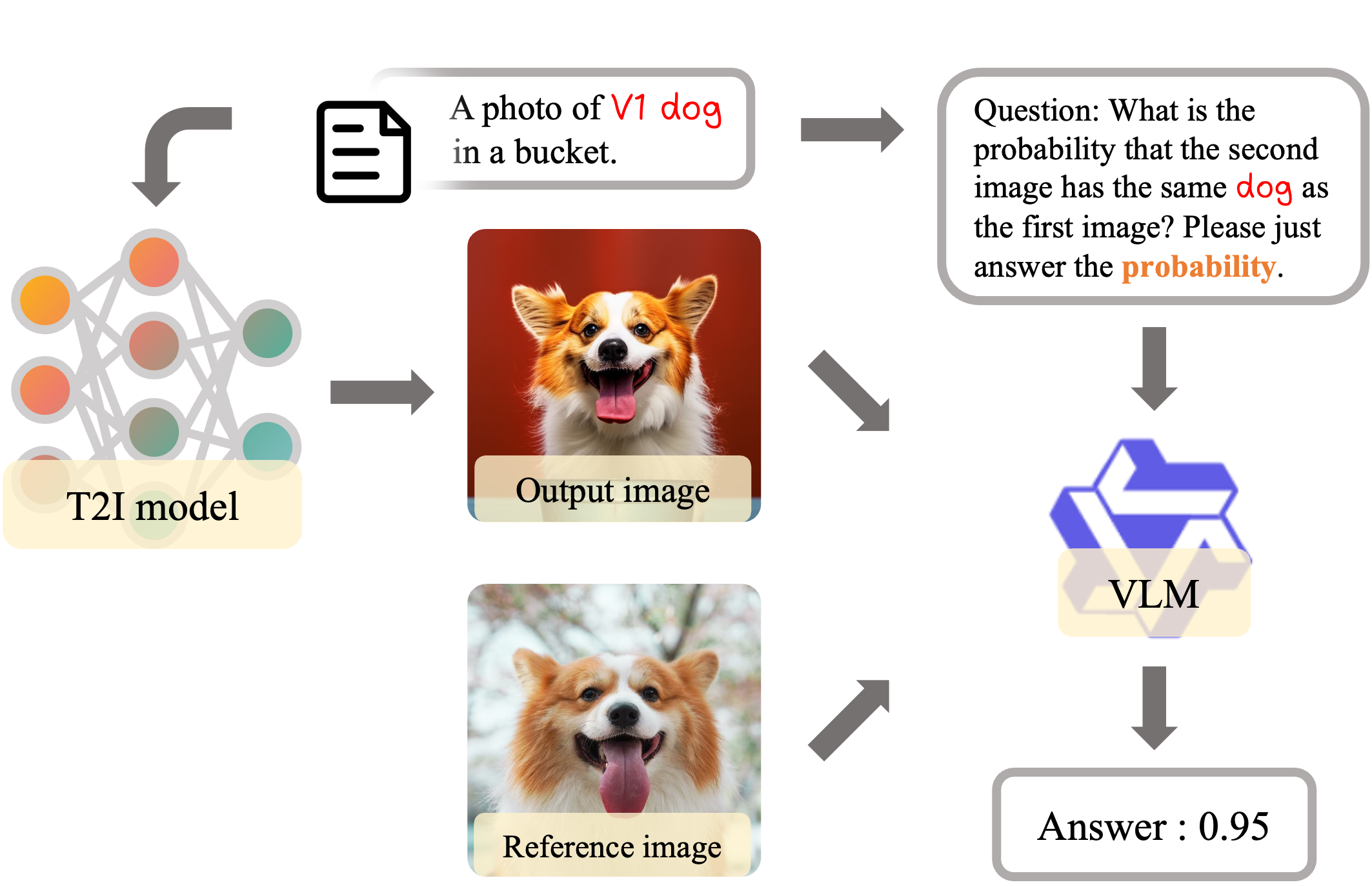}
    \caption{Evaluation pipeline of the unique personalized item similarity by VQA for Item customization tasks.}
    \label{fig:item_sim}
\end{figure}

\noindent\ding{182} \textit{Item Customization}
For each item customization task, we evaluate the model’s ability to generate each fine-grained concept independently. Specifically, for each unique item, we prompt the post-trained model to produce a test image and use the original training-set image as a reference. We then convert each test prompt into a corresponding question using the template in \Tab~\ref{tab:item_sim_template}. Finally, a VLM assesses the similarity between the generated image and its reference to produce the score $\mathrm{Sim}(i,k)=\{\mathrm{score}\}\times 100\%$, representing the similarity score of the $i$-th question in the $k$-th unique personalized item. The pipeline is illustrated in \Fig~\ref{fig:item_sim}.
The \textbf{Unique-Sim} metric is calculated by the average of all unique personalized items:
\begin{equation}
{\textbf{Unique-Sim}} = \frac{1}{K} \sum_{k=1}^{K}\frac{1}{N_k} \sum_{i=1}^{N_k} \mathrm{Sim}(i, k),
\end{equation}
where $K$ is the number of item customization tasks, $N_k$ is the number of question-image pairs corresponding to the $k$-th unique personalized item.
\begin{table}[h]
\centering
\caption{The corresponding templates of prompt words and questions. When calculating unique accuracy, a corresponding question and reference image pair is generated for each personalized prompt. When calculating class similarity, four questions and reference image pairs are generated for each class prompt.}
\begin{tabular}{ccp{8cm}}  
\toprule
\textbf{Class} & \textbf{Unique} & \textbf{Question Template} \\
\midrule
dog & V1 dog & What is the probability that the second image has the same \textit{dog} as the first image? Please just answer the probability. \\
\midrule
dog & V2 dog & What is the probability that the second image has the same \textit{dog} as the first image? Please just answer the probability. \\
\midrule
cat & V3 cat & What is the probability that the second image has the same \textit{cat} as the first image? Please just answer the probability. \\
\midrule
sneaker & V4 sneaker & What is the probability that the second image has the same \textit{sneaker} as the first image? Please just answer the probability. \\
\bottomrule
\end{tabular}
\label{tab:item_sim_template}
\end{table}

\noindent\ding{183} \textit{Domain Enhancement} \href{https://github.com/tgxs002/HPSv2}{Human Preference Score (HPS)}~\cite{DBLP:conf/iccv/WuSZZL23} is an automated evaluation metric trained to predict human judgments on T2I outputs by fine-tuning a CLIP model on the large-scale Human Preference Dataset. We employ HPS to evaluate the Body and the Nature domain for the alignment of the generated images with human aesthetics, respectively as \textbf{Body-HPS} and \textbf{Nature-HPS}.

\textbf{Forget Measure}
Beyond measuring degradation of pretrained capabilities relative to the base model, we also quantify forgetting in downstream performance dynamics during continual post‑training and class concept forgetting in specific item customization tasks: 

\noindent\ding{182} \textit{Backward Transfer}
For both Item Customization and Domain Enhancement, we compute \textit{backward transfer} on their respective downstream metrics to evaluate the knowledge stability across sequential tasks. Backward transfer is the relative influence of learning the $k$-th task on all old tasks, defined as follows:
\begin{equation}
\textbf{Forget} = \frac{1}{K-1}\sum_{k=1}^{K-1}\mathrm{BWT}_k,\quad\text{BWT}_k = \frac{a_{k,k} - a_{K,k}}{a_{k,k}},
\end{equation}
where $a_{k,j}$ is the evaluation metric for the $j$-th task after the $k$-th round of training. Negative values indicate performance degradation on earlier tasks. By substituting the task-sequence Unique-Sim and HPS values into $a$, we can get the \textbf{Unique-Forget} and \textbf{Domain-Forget} metrics, respectively.
    
\noindent\ding{183} \textit{Class Concept Forgetting}   
When learning personalized concepts in item customization tasks, we evaluate the forgetting of their corresponding base classes. We generate images for non-personalized prompts (e.g., "a dog...") and evaluate their similarity with all learned unique personalized items (e.g., "V1 dog..."). Combined with the designed question template, each test prompt is converted into multiple corresponding personalized similarity questions. The question template is shown in \Tab~\ref{tab:item_sim_template}. The VLM model is used to test the similarity of the generated image with the reference image in the personalized concept and obtain a score $\mathrm{Sim}(i_k,j)=\{\mathrm{score}\}\times 100\%$ as in Unique-Sim. 
We use $\mathrm{Sim}(i_k,j)$ to represent the similarity score of images generated by the $i$-th prompt of class $k$ with the $j$-th unique item. Then we need to compute \textbf{Class-Sim} as BWT by evaluating each current class's prompts with all old classes:
\begin{equation}
{\textbf{Class-Sim}} = \frac{1}{K}\sum_{k=1}^{K}\frac{1}{N_k} \sum_{i=1}^{N_k}\frac{1}{K} \sum_{j=1}^{K} \mathrm{Sim}(i_k, j)
\end{equation}
where $N_k$ is the number of non-personalized prompts in class $k$.

\begin{figure}[htb]
    \centering
    \includegraphics[width=\linewidth]{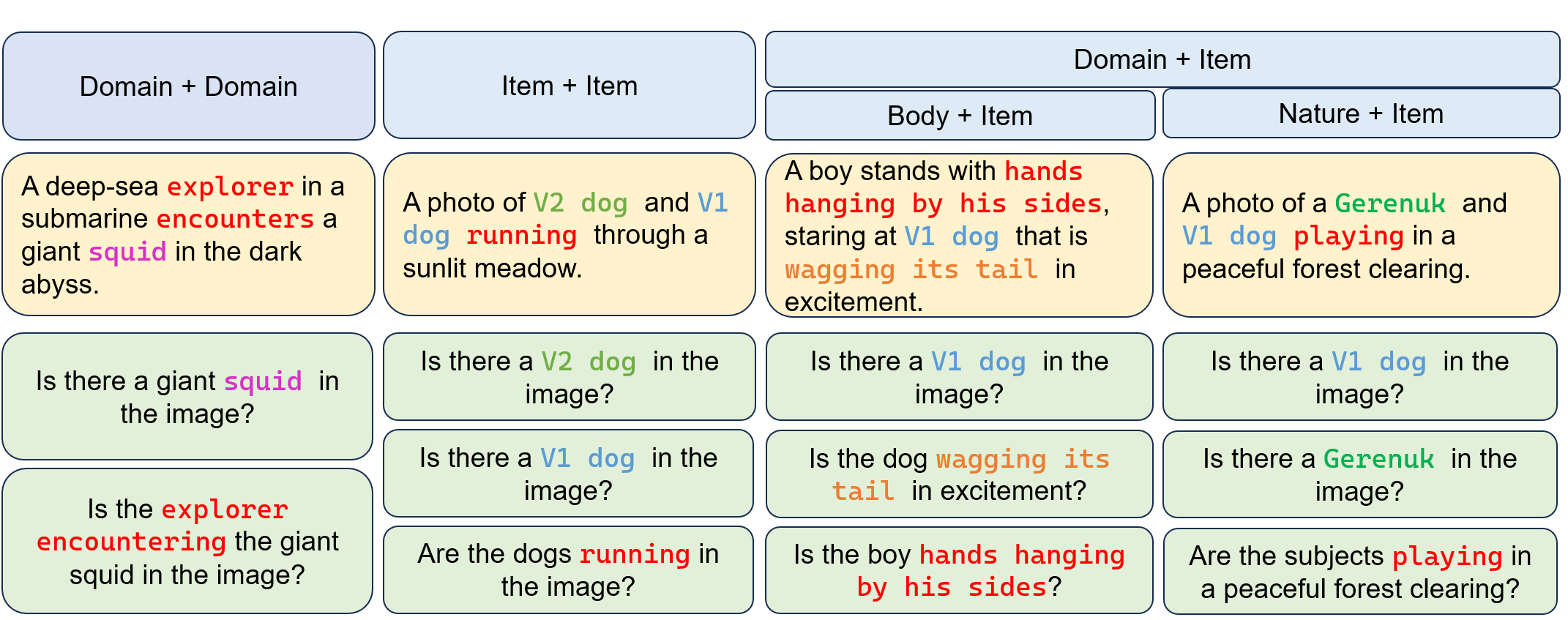}
    \caption{Example decomposition of four cross-task prompts into questions across different combination types. Colored highlights in each prompt and question indicate the key objects and actions under evaluation.}
    \label{fig:cross template}
\end{figure}

\textbf{Cross-task Generalization}
To evaluate the post-trained model's ability to recombine concepts from different tasks, we generate novel, compositional prompts as described in \Sec~\ref{sec:data} and measure how accurately the continually learned model renders them. This evaluation follows a three-step VQA-based pipeline (see \Fig~\ref{fig:cross_eval}):


\noindent\ding{182} \textit{Prompt Decomposition}
We use an LLM to break each compositional test prompt into 2–4 simpler questions that collectively cover all relevant objects and their interactions (e.g., “Are the dogs running in the image?”). We ensure these sub-questions comprehensively probe both individual elements (vertical objects, personalized instances) and their relational actions. Example templates and generated Q\&A pairs for different test sets are shown in \Fig~\ref{fig:cross template}.

\noindent\ding{183} \textit{VQA Formatting} 
Each LLM-generated question $t$ is formatted into a VQA-compatible query $q(t)$. For generic scenes without specialized objects, we simply append “Please answer yes or no.”
\[
\begin{aligned}
\mathbf{t} &= 
\begin{minipage}[t]{12cm}
Are the dogs running in the image?
\end{minipage} \\
\mathbf{q(t)} &= 
\begin{minipage}[t]{12cm}
"Are the dogs running in the image?" Please just answer yes or no.
\end{minipage}
\end{aligned}
\]

For questions involving natural or personalized objects, we apply object-specific templates (\Tab~\ref{tab:cross question}).

\begin{table}[htb]
\centering
\caption{Example templates for VQA prompts that require reference images: each question is converted into a formatted question for the VLM, illustrating how personalized items and natural species are described and queried in a two-image comparison.}
\begin{tabular}{ccp{8cm}}  
\toprule
\textbf{Task} & \textbf{Concept} & \textbf{Formatted Question} \\
\midrule
\multirow{16}{*}{Item} & V1 dog & "The image of the V1 dog is image 1, please identify the breed of this dog. For image 2, is there a dog of the same breed and similar appearance? Please just answer yes or no." \\

 & V2 dog & "The image of the V2 dog is image 1, please identify the breed of this dog. For image 2, is there a dog of the same breed and similar appearance? Please just answer yes or no."  \\

 & V3 cat & "The image of the V3 cat is image 1, please identify the breed of this cat. For image 2, is there a cat of the same breed and similar appearance? Please just answer yes or no."\\

 & V4 sneaker & "The image of the V4 sneaker is image 1, please identify the style of this sneaker. For image 2, is there a sneaker of the same style and similar appearance? Please just answer yes or no." \\
\midrule
\multirow{13}{*}{Nature} & pomelo & "The image of the pomelo is image 1. Image 2 is a part of the pomelo. For image 3" $+$ question $t$ \\

 & Spix's macaw & "The image of Spix's macaw is image 1, have over 30 percent blue feathers. For image 2," $+$ question $t$ \\

 & Squid & "The image of Squid is image 1. Note that Squids have a distinct elongated body and tentacles, and should not be confused with Octopuses, which have a more rounded body and eight arms without distinct tentacles. For image 2," $+$ question $t$ \\

 & Quokka & "Only animals that are many similar to the one in image 1 will be considered Quokka.For image 2," $+$ question $t$  \\

 & Gerenuk & "Only animals that are many similar to the one in image 1 will be considered Gerenuk.For image 2," $+$ question $t$  \\
\bottomrule
\end{tabular}
\label{tab:cross question}
\end{table}


\noindent\ding{184} \textit{Answer Scoring}
The visual-language model (VLM) processes each image–question pair $(x,q(t))$ and returns “yes” or “no.” We assign a score of 1 for “yes” and 0 otherwise. The overall cross-task score for a test set is the fraction of “yes” responses across all $N$ image–question pairs:
\begin{equation}
\text{score}(x, q(t)) = 
\begin{cases} 
1&\text{, answer= "yes"} \\
0&\text{, otherwise}
\end{cases}
\end{equation}
Finally, the test score of the cross-task test set is defined as the proportion of “yes” answers among all question-answer pairs in the test set:
\begin{equation}
\text{Cross} = \frac{1}{N} \sum_{i=1}^N \text{score}(x_i, q(t_i))
\end{equation}
We denote cross-task generalization metrics for different task combinations as \textbf{Item$+$Item}, \textbf{Item$+$Domain}, and \textbf{Domain$+$Domain}, respectively.


\textbf{Remark}
We use the open-source LLMs DeepSeek V3~\cite{deepseekai2025deepseekv3technicalreport} and DeepSeek R1~\cite{deepseekai2025deepseekr1incentivizingreasoningcapability}. Our VQA pipeline employs Qwen2.5-7B-Instruct~\cite{qwen2025qwen25technicalreport} as the VLM. We acknowledge that more advanced—and potentially more accurate—models like GPT-4V~\cite{yang2023dawnlmmspreliminaryexplorations} exist for evaluation, but we provide a minimal, fully reproducible setup with open-source models better suited for benchmarking. We also retain interfaces that allow seamless integration of more advanced evaluators as the benchmark evolves.

\section{Detailed Continual Post-training Baselines}\label{sec:appendix baselines}


\begin{itemize}[left=0pt,itemsep=0pt,topsep=0pt]

\item \textbf{Base} employs the pretrained model without further continual post-training, establishing a baseline on general generative capabilities and downstream tasks. 
\item \textbf{Joint}~\cite{DBLP:conf/icassp/WuTPK23} jointly trains the model on all task data, characterizing the upper bound of performance in sequential learning.
\item \textbf{SeqFT}~\cite{zhang2024slcaunleashpowersequential,chen2025blip3ofamilyfullyopen} sequentially fine-tunes the model on each task with all parameters updated in task order. The model is optimized exclusively for the current task without preserving pretrained knowledge or retaining performance on earlier tasks. 

\item \textbf{Replay}~\cite{chaudhry2019tinyepisodicmemoriescontinual} maintains a small memory buffer that stores samples to replay prior knowledge. We store 10\% of each completed task’s image–text pairs in a small memory buffer and mix them with new‑task data during subsequent fine‑tuning. For item datasets with fewer than 10 examples, we ensure at least one sample is retained for replay.
\item \textbf{$\ell_2$-norm}~\cite{DBLP:conf/icml/ZhaoW0L24} adds an $\ell_2$‑norm penalty on the change from the previous task’s final parameters. Concretely, when training on task $i$, the loss becomes $\mathcal L_i=\mathcal L_{i}^{\mathrm{new}} + \lambda \Omega_i(\theta_i,\theta_{\mathrm{old}})$, where $\mathcal L_i^{\mathrm{new}}$ is the standard loss on the new task, $\theta_{\mathrm{old}}$ are the frozen parameters from previous tasks, $\Omega_i$ is the regularization function, and $\lambda$ controls its strength. Formally, the regularization term of $\ell_2$‑norm is $\Omega_{\ell_2}=\lVert\theta_i-\theta_{i-1} \rVert_2$. This term discourages large deviations from the starting values at the beginning of task $i$, thereby limiting drastic parameter shifts when learning the new task.
\item \textbf{EWC}~\cite{Kirkpatrick_2017} is built upon the $\ell_2$‑Norm baseline by weighting each parameter’s penalty according to its estimated importance to previous tasks. Let $F_{k}$ be the Fisher Information Matrix (FIM)~\cite{liao2018approximatefisherinformationmatrix} computed after task $k$. We form a diagonal approximation on all old tasks $\hat{F}_{1: i-1}=\sum_{k=1}^{i-1}\mathrm{diag}(F_k)$. When training on task $i$, the regularization term is $\Omega_{\mathrm{EWC}}=\left(\theta_i-\theta_{i-1}\right)^{\top} \hat{F}_{1: i-1}\left(\theta_i-\theta_{i-1}\right)$. Parameters with higher Fisher scores incur a larger penalty for deviation, thereby more effectively preserving those weights most critical to earlier tasks.



\item \textbf{HFT}~\cite{DBLP:journals/corr/abs-2404-18466} randomly splits parameters into two groups before each new task, i.e., $\theta_k=\{\vartheta_k, \psi_k\}$. One half (50\%) $\vartheta_k$ is updated on the new task, $\vartheta^t_k \leftarrow \vartheta^{t-1}_k-\eta \nabla_{\vartheta_k} \mathcal{L}\left(\theta^{t-1}_k\right)$, while the other half $\psi_k$ remains frozen to preserve prior knowledge, $\psi_k^t \leftarrow \psi^t_{k-1}$. During each task of continual post‑training, only the active group is tuned, achieving a dynamic balance between learning new concepts and retaining old ones.
\item \textbf{MoFO}~\cite{chen2025mofomomentumfilteredoptimizermitigating} leverages the momentum terms from the Adam optimizer~\cite{DBLP:journals/corr/KingmaB14} to approximate parameter importance. To keep computation efficient, MoFO first groups parameters by their natural components (e.g., weight matrices versus bias vectors).
After each backward pass, parameters are ranked by the absolute value of their momentum. MoFO updates only parameters with the largest $\alpha$\% momentum in each partition for critical directions, and the rest remain frozen. By focusing updates on these “high-momentum” directions, MoFO achieves a sparse, adaptive fine-tuning that accelerates learning of new tasks without destabilizing performance on previously learned tasks.


\item \textbf{SeqLoRA}~\cite{DBLP:conf/naacl/DevlinCLT19} shares a single LoRA adapter across all tasks. The LoRA adapter factorizes the update as $\Delta W \approx BA$, where $B\in\mathbb{R}^{d\times r}$ and $A\in\mathbb{R}^{r\times d}$ are low‑rank matrices with $r \ll d)$. During post-training on each task, the original weights remain frozen and only $A$ and $B$ are learned. This approach is simple and efficient, but may suffer from interference between tasks.
\item \textbf{IncLoRA}~\cite{DBLP:conf/emnlp/WangCGXBZZGH23} allocates a fresh, independent LoRA adapter ($B_i, A_i$) for each new task $i$. 
The model's effective weight after $i$ tasks for inference is $W_i=W_0+\sum_{k=1}^i B_k A_k$
By assigning each task its own low‑rank subspace, it enforces strict task isolation at the cost of linearly increasing the number of parameters.
\item \textbf{O-LoRA}~\cite{DBLP:conf/emnlp/WangCGXBZZGH23} 
extends IncLoRA by imposing orthogonality across task adapters. When training the $i$-th adapter, it minimizes the task loss subject to $L_{\text {O-LoRA}}=\lambda \sum_{j=1}^{i-1}\lVert B_j^\top B_i\rVert^2$, ensuring each task’s low-rank subspace remains mutually orthogonal. $\lambda$ is the coefficient to control regularization strength. This further reduces parameter conflict across tasks and enhances knowledge separation.
\item \textbf{C-LoRA}~\cite{DBLP:journals/tmlr/SmithHZHKSJ24} adds a self‑regularization term that penalizes deviations between the LoRA update for the new task and the adapters learned for previous tasks. The addition loss term for task $i$ is
$\mathcal{L}_{\text {C-LoRA}}=\lambda\left\|\left[\sum_{j=1}^{t-1} \boldsymbol{A}_{j} \boldsymbol{B}_{j}\right] \odot \boldsymbol{A}_i \boldsymbol{B}_i\right\|_2^2$, where $\lambda$ balances adaptation to the new task with consistency to prior updates.
By encouraging consistency with prior adapters, C‑LoRA strikes a balance between retaining old knowledge and adapting to new tasks.
\end{itemize}

\section{Detailed Training Implementation}\label{sec:appendix train}

\textbf{Sequential Item Customization} 
We build on the \href{https://github.com/huggingface/diffusers/tree/main/examples/dreambooth}{Diffusers DreamBooth example}, integrating DeepSpeed~\cite{rajbhandari2020zeromemoryoptimizationstraining} Stage 2 for memory-efficient training. All methods fine-tune using the following shared settings unless noted otherwise:
\begin{itemize}[left=0pt,itemsep=0pt,topsep=0pt]
    \item Optimizer: AdamW~\cite{DBLP:conf/iclr/LoshchilovH19} with learning rate of $5\times10^{-6}$, weight decay $1\times10^{-2}$, gradient clipping at $1.0$.
    \item Batch size $=4$.
    \item Scheduler: constant learning rate.
    \item Training Steps: 500 for each item.
\end{itemize}
For each task, we use 500 prior class images generated by the base model to prevent overfitting on each personalized concept, with a prior regularization coefficient of $0.02$.

Baseline-specific configurations: 
\begin{itemize}[left=0pt,itemsep=0pt,topsep=0pt]
    \item Joint: train 2000 steps on all item datasets.
    \item LoRA Variants (SeqLoRA, IncLoRA, O-LoRA, C-LoRA): rank $= 16$, LoRA $\alpha=32$.
    \item O-LoRA: orthogonality penalty $\lambda=1\times10^{-1}$.
    \item C-LoRA: self-regularization $\lambda=1\times10^6$.
    \item $\ell_2$-norm: regularization coefficient $\lambda=1\times10^{-3}$.
    \item EWC: regularization coefficient $\lambda=1\times10^{-4}$. 
    \item HFT: freeze half of each layer’s parameters (freeze ratio = $0.5$).
    \item MoFO: partition at the parameter level, updating only the top 50\% by momentum ($\alpha = 0.5$) and build upon Adam~\cite{DBLP:journals/corr/KingmaB14}.
\end{itemize}

\textbf{Sequential Domain Enhancement} 
We build on the \href{https://github.com/PixArt-alpha/PixArt-alpha}{PixArt-$\alpha$} training pipeline, integrating DeepSpeed Stage 2 for efficient memory usage. All methods share these base settings unless specified otherwise:
\begin{itemize}[left=0pt,itemsep=0pt,topsep=0pt]
    \item Optimizer: AdamW~\cite{DBLP:conf/iclr/LoshchilovH19} with learning rate of $5\times10^{-6}$, weight decay $1\times10^{-2}$, gradient clipping at $1.0$.
    \item Batch size $=256$.
    \item Scheduler: constant learning rate.
    \item Training Steps: 3000 for each domain.
\end{itemize}
Baseline-specific configurations: 
\begin{itemize}[left=0pt,itemsep=0pt,topsep=0pt]
    \item Joint: train 48000 steps on all domain datasets.
    \item LoRA Variants (SeqLoRA, IncLoRA, O-LoRA, C-LoRA): rank $= 16$, LoRA $\alpha=32$.
    \item O-LoRA: orthogonality penalty $\lambda=1\times10^{-1}$.
    \item C-LoRA: self-regularization $\lambda=1\times10^6$.
    \item $\ell_2$-norm: regularization coefficient $\lambda=1\times10^{-3}$.
    \item EWC: regularization coefficient $\lambda=1\times10^{-4}$. 
    \item HFT: freeze half of each layer’s parameters (freeze ratio = $0.5$).
    \item MoFO: partition at the parameter level, updating only the top 50\% by momentum ($\alpha = 0.5$) and build upon Adam~\cite{DBLP:journals/corr/KingmaB14}.
\end{itemize}

\section{Additonal Experiment Results}\label{sec:appendix experiment}

\begingroup
\renewcommand{\arraystretch}{0.9}
\begin{table}[t]
  \caption{Performance of continual post-training methods for the \textit{sequential item-domain adaptation} task of \textit{Order~2} using \textit{SD v1.4}. $\uparrow$: higher is better. $\downarrow$: lower is better. “I” and “D” denote Item and Domain, respectively, with combinations indicating cross-task generalization evaluations. Excluding \textit{Base} and \textit{Joint}, the best result is shown in \textbf{bold} and the second-best is \underline{underlined}.}
  \vspace{3pt}
  \centering
  \resizebox{\textwidth}{!}{
    \begin{tabular}{>{\centering\arraybackslash}c|lccccccccc}
      \toprule
      \multirow{2}{*}{\textbf{Order~2}} &
      \multirow{2}{*}{\textbf{Method}} &
      \multicolumn{2}{c}{\textbf{Pretrain}} &
      \multicolumn{1}{c}{\textbf{Item}} &
      \multicolumn{2}{c}{\textbf{Domain}} &
      \multicolumn{3}{c}{\textbf{Cross}} &
      \multicolumn{1}{c}{\textbf{Forget}} \\
      \cmidrule(lr){3-4}\cmidrule(lr){5-5}\cmidrule(lr){6-7}\cmidrule(lr){8-10}\cmidrule(lr){11-11}
      & & \textbf{FID} $\downarrow$ & \textbf{Comp} $\uparrow$ & \textbf{Unique-Sim} $\uparrow$ & \textbf{Body-HPS} $\uparrow$ & \textbf{Nature-HPS} $\uparrow$ &
      \textbf{I+I} $\uparrow$ & \textbf{I+D} $\uparrow$ & \textbf{D+D} $\uparrow$ & \textbf{Class-Sim} $\downarrow$ \\
      \midrule
      \multirow{12}{*}{\makecell[c]{“Nature”\\$\downarrow$\\“Body”\\$\downarrow$\\“V1 dog”\\$\downarrow$\\“V2 dog”\\$\downarrow$\\“V3 cat”\\$\downarrow$\\“V4 sneaker”}} &
      \textit{Base}   &  9.9275 & 0.2901 & 0.0000 & 0.2118 & 0.2229 & 0.1806 & 0.2224 & 0.1493 & 0.0013 \\
      & \textit{Joint} & 22.7432 & 0.3097 & 0.1225 & 0.2968 & 0.2851 & 0.2028 & 0.3020 & 0.2985 & 0.0293 \\
      & SeqFT          & 19.0929 & 0.3043 & 0.3450 & \textbf{0.2919} & 0.2598 & \textbf{0.3444} & 0.2632 & 0.2289 & \underline{0.0238} \\
      \cmidrule(lr){2-11}
      & SeqLoRA        & 16.5584 & 0.2805 & 0.3025 & 0.2519 & 0.2422 & 0.3111 & 0.2918 & 0.1940 & 0.0788 \\
      & IncLoRA        & 17.7793 & 0.2766 & 0.2675 & 0.2473 & 0.2502 & 0.3306 & 0.3061 & 0.1791 & 0.0850 \\
      & O-LoRA         & \underline{14.1877} & 0.2727 & 0.2700 & 0.2425 & 0.2553 & 0.2778 & 0.2958 & 0.1244 & 0.0458 \\
      & C-LoRA         & \textbf{14.1097} & 0.2804 & 0.2975 & 0.2465 & 0.2564 & 0.2778 & 0.2754 & 0.1443 & 0.0550 \\
      \cmidrule(lr){2-11}
      & $\ell_2$-norm  & 14.7921 & 0.2992 & 0.2300 & 0.2680 & 0.2577 & 0.2722 & 0.3081 & \underline{0.2587} & 0.0475 \\
      & EWC            & 19.3321 & 0.2883 & \textbf{0.5050} & 0.2794 & \underline{0.2691} & 0.2917 & 0.2959 & 0.1990 & 0.0543 \\
      \cmidrule(lr){2-11}
      & HFT            & 16.6841 & \textbf{0.3136} & \underline{0.3525} & 0.2901 & 0.2633 & 0.3111 & \underline{0.3121} & \textbf{0.2786} & \textbf{0.0093} \\
      & MoFO           & 17.0268 & 0.3053 & 0.2600 & \underline{0.2907} & 0.2650 & 0.2917 & 0.2939 & 0.2239 & 0.0418 \\
      & Replay         & 17.8449 & \underline{0.3084} & 0.3450 & 0.2884 & \textbf{0.2796} & \underline{0.3333} & \textbf{0.3122} & 0.2189 & 0.0668 \\
      \bottomrule
    \end{tabular}
  }
  \label{tab:sdv14_item_domain_order2}
  \vspace*{-2mm}
\end{table}
\endgroup

\subsection{Continue Post-training for Sequential Item-Domain Adaptation on SD v1.4}

In addition to PixArt-$\alpha$ based on DiT~\cite{DBLP:conf/iccv/PeeblesX23} used in \Sec~\ref{sec:experiments}, we also experiment with \textbf{Stable Diffusion v1.4} (\textbf{SD~v1.4})~\cite{DBLP:conf/cvpr/RombachBLEO22}, a U-Net–based model, as the base model. We apply the same Order~2 of Sequential Item-Domain Adaptation task (\Tab~\ref{tab:pixart_alpha_item_domain_order12}) to evaluate various continual post-training methods. The results are shown in \Tab~\ref{tab:sdv14_item_domain_order2}. Overall, the findings mirror our key takeaways. A detailed analysis follows:

\textbf{FID} Since SD~v1.4 exhibits stronger pretrained generative capabilities, all continual post-training methods show a distribution shift–induced quality drop on the new datasets.

\textbf{Text–Image Alignment (Comp)} Nearly every method improves alignment, except the LoRA variants and EWC. 

\textbf{Item} Joint suffers from domain data bias and remains weak on item Unique-Sim. EWC achieves the best item metrics, and notably, Replay also successfully learns item concepts on SD~v1.4 (unlike on PixArt-$\alpha$), indicating that Replay’s effectiveness varies across architectures.

\textbf{Domain} Joint continues to set the upper bound. Most methods exhibit forgetting on the first Nature domain; Replay best updates and preserves Nature domain knowledge, followed closely by EWC.

\textbf{Cross-Task Generalization} Joint excels only on Domain$+$Domain composition and underperforms on item-domain mixed generalization. All LoRA variants struggle, whereas HFT emerges as the strongest overall, underscoring the effective and efficient solution of parameter and feature reuse for knowledge fusion.

\subsection{Visualization of Cross-task Generalization Results}

\begin{figure}[htb]
    \centering
    \includegraphics[width=0.65\linewidth]{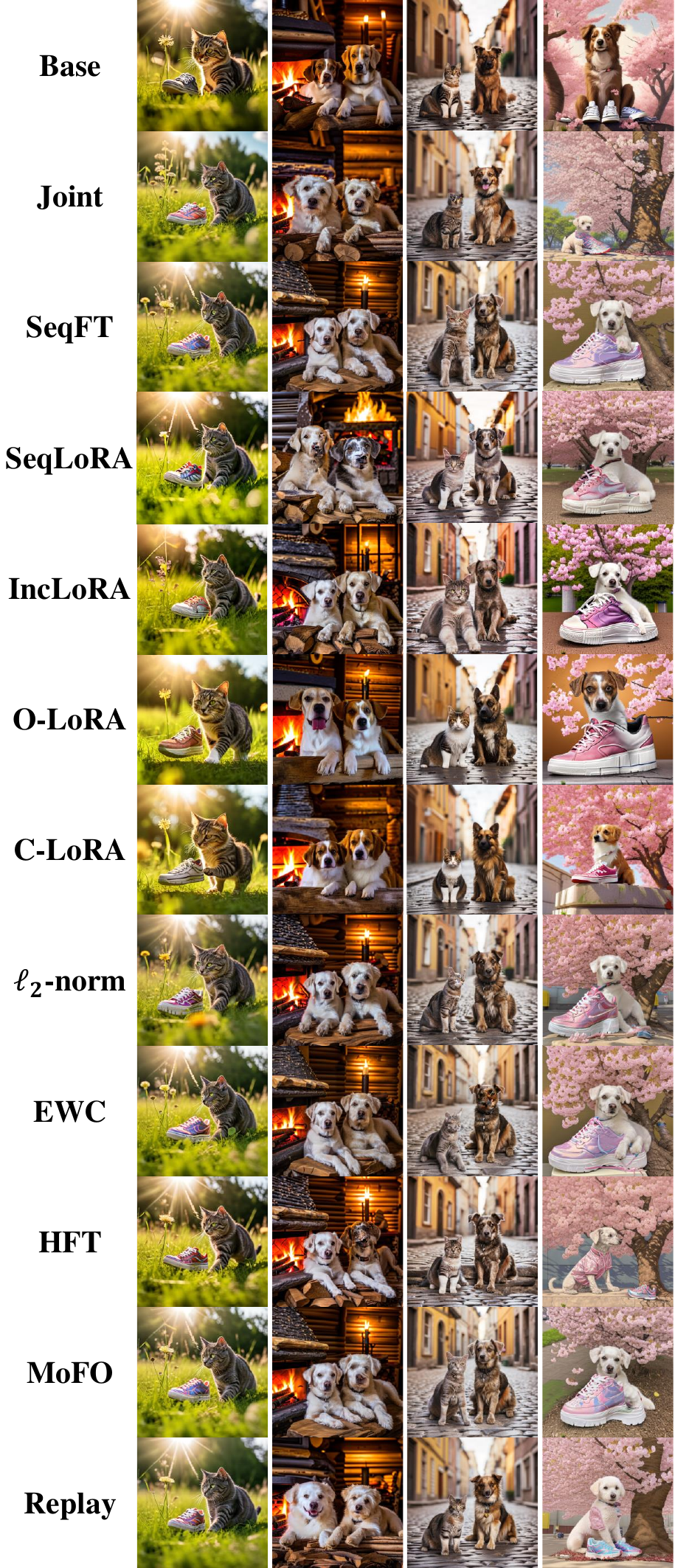}
    \caption{Results on the Item$+$Item cross-task test set by the models of sequential item customization in \Tab~\ref{tab:pixart_alpha_item_and_domain}. Prompts of each column: (1)A photo of V3 cat playing with V4 sneaker in a sunlit meadow; (2)A photo of V1 dog and V2 dog relaxing near a crackling fireplace in a log cabin; (3)A photo of V1 dog and V3 cat sitting together on a cobblestone street in an old town; (4)A depiction of V1 dog sitting with V4 sneaker under a cherry blossom tree in full bloom.}
    \label{fig:item_item_visual_result}
\end{figure}

\begin{figure}[htb]
    \centering
    \includegraphics[width=0.70\linewidth]{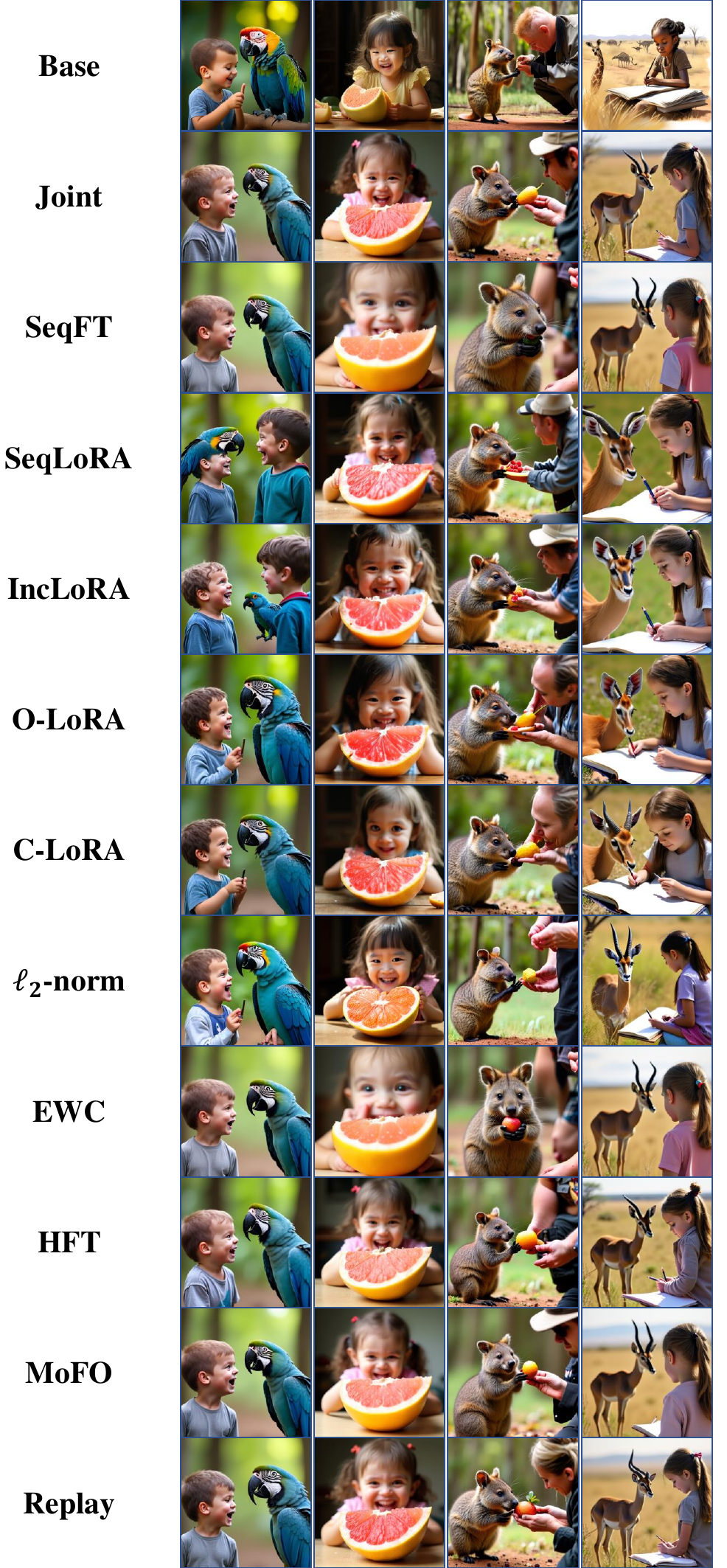}
    \caption{Results on the Domain$+$Domain cross-task test set by the models of sequential domain enhancement in  \Tab~\ref{tab:pixart_alpha_item_and_domain}. Prompts of each column: (1)A little boy joyfully watches as a Spix’s macaw mimics his whistling sounds; (2)A little girl struggles to peel a pomelo, her face lighting up as she finally separates the segments; (3)A park ranger gently feeds a quokka a small piece of fruit; (4)A girl sketches a gerenuk in her wildlife observation journal.}
    \label{fig:domain_domain_visual_result}
\end{figure}

\begin{figure}[htb]
    \centering
    \includegraphics[width=0.70\linewidth]{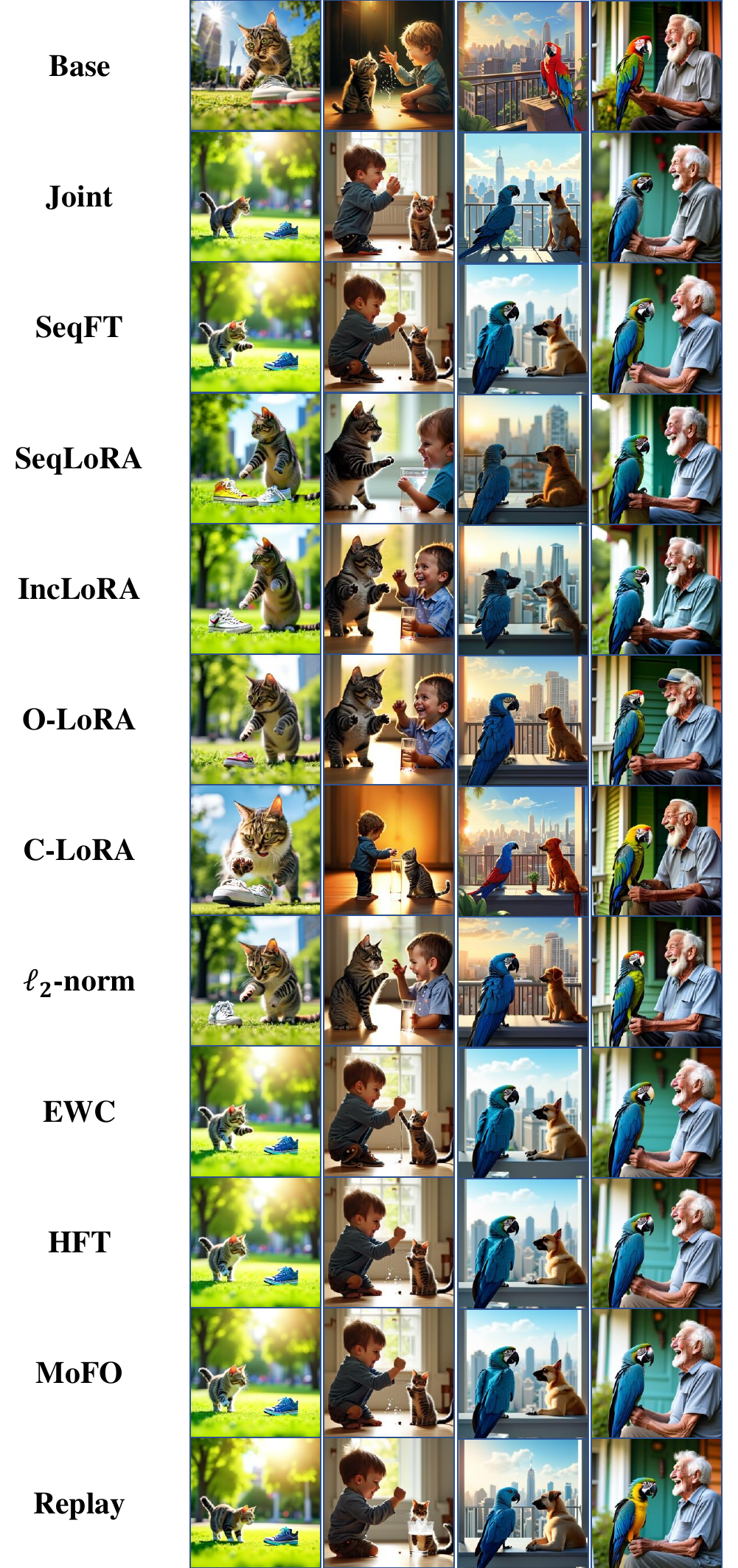}
    \caption{Results on the cross-task test set by the models of sequential item-domain adaptation Order~1 in \Tab~\ref{tab:pixart_alpha_item_domain_order12}. Prompts of each column: (1)Item$+$Item: A scene of V3 cat playing with V4 sneaker in a city park on a bright summer day; (2)Item$+$Body: A little boy makes a fist and shakes it playfully at a mischievous V3 cat that has just knocked over; (3)Item$+$Nature: A depiction of a Spix's Macaw and V2 dog relaxing on a balcony overlooking a modern cityscape; (4)Domain$+$Domain: An elderly man on his porch talks to his pet Spix’s macaw, which responds with cheerful squawks.}
    \label{fig:item-domain-order1_visual_result}
\end{figure}

\begin{figure}[htb]
    \centering
    \includegraphics[width=0.70\linewidth]{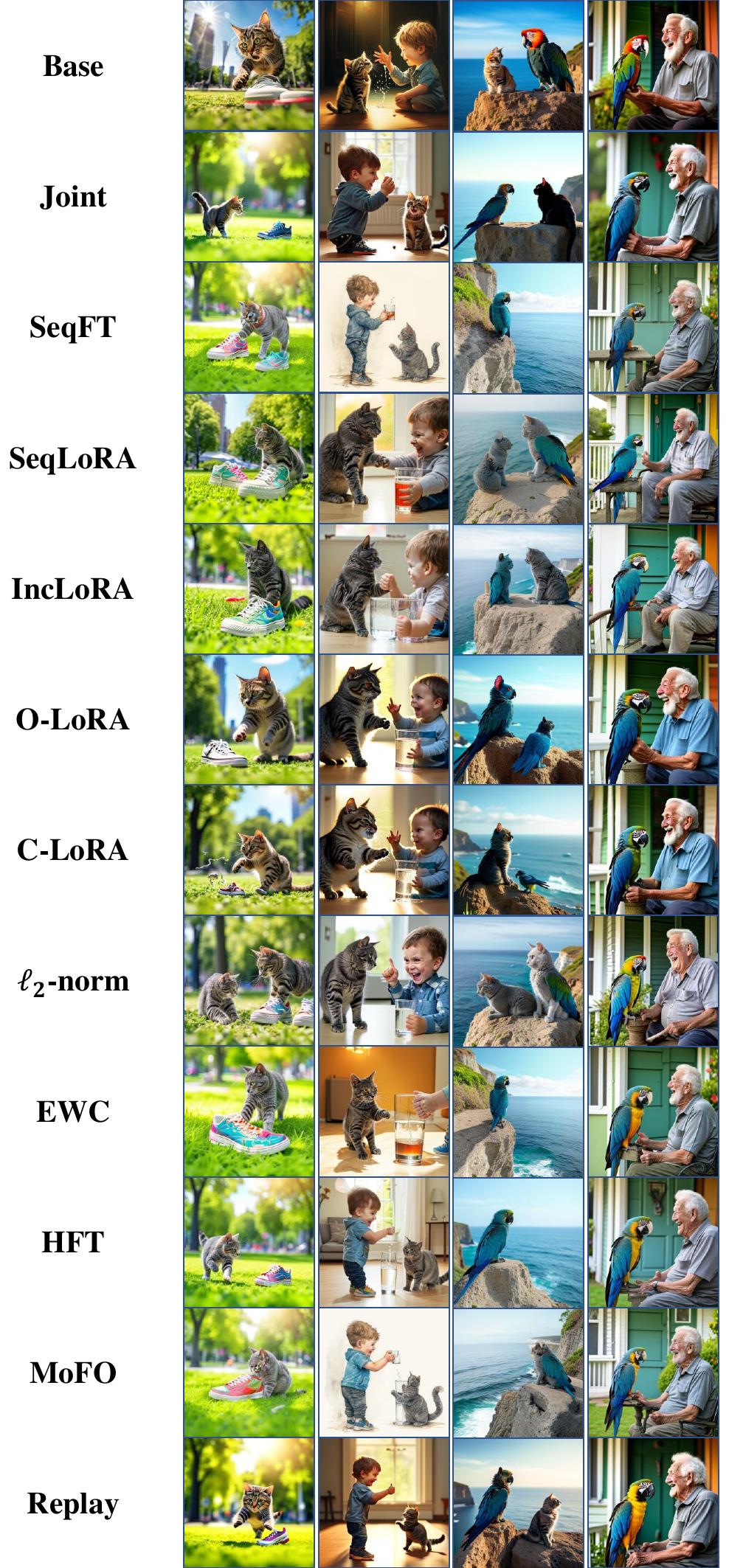}
    \caption{Results on the cross-task test set by the models of sequential item-domain adaptation Order~2 in \Tab~\ref{tab:pixart_alpha_item_domain_order12}. Prompts of each column: (1)Item$+$Item: A scene of V3 cat playing with V4 sneaker in a city park on a bright summer day; (2)Item$+$Body: A little boy makes a fist and shakes it playfully at a mischievous V3 cat that has just knocked over; (3)Item$+$Nature: A photo of a Spix's Macaw and V3 cat perched together on a cliff overlooking the ocean; (4)Domain$+$Domain: An elderly man on his porch talks to his pet Spix’s macaw, which responds with cheerful squawks.}
    \label{fig:item-domain-order2_visual_result}
\end{figure}

{\Fig}~\ref{fig:item_item_visual_result},\ref{fig:domain_domain_visual_result},\ref{fig:item-domain-order1_visual_result},\ref{fig:item-domain-order2_visual_result} present example cross-task generalization test images generated by the models across the three sequential task settings.



\end{document}